\documentclass[letterpaper, 10 pt, conference]{ieeeconf}  %

\IEEEoverridecommandlockouts                              %

\usepackage{times}

\usepackage{cite}
\usepackage{multicol}
\usepackage[bookmarks=true]{hyperref}
\usepackage[utf8]{inputenc}

\usepackage{xcolor}
\usepackage{etoolbox}
\usepackage[linesnumbered,vlined,ruled,commentsnumbered]{algorithm2e}
\SetKwRepeat{Do}{do}{while}

\SetVlineSkip{0.2em}

\SetInd{0.2em}{0.7em}

\DontPrintSemicolon{}

\SetKwIF{If}{ElseIf}{Else}{if}{}{else if}{else}{endif}

\SetAlgoSkip{}

\SetAlCapHSkip{0cm}

\SetAlgoCaptionLayout{small}
\SetAlFnt{\small}

\SetArgSty{textnormal}

\SetKwComment{Comment}{$\triangleright$\ }{}

\makeatletter
\patchcmd\algocf@Vline{\vrule}{\vrule \kern-0.4pt}{}{}
\patchcmd\algocf@Vsline{\vrule}{\vrule \kern-0.4pt}{}{}
\makeatother
\usepackage{tikz}

\usepackage{tikzscale}

\usepackage{pgfplots}

\usepackage{pgfplotstable}

\usetikzlibrary{calc}

\usetikzlibrary{positioning}

\usepgfplotslibrary{statistics}

\usepgfplotslibrary{fillbetween}

\pgfplotsset{
  compat=1.15,
  mps basic/.style={
    xlabel near ticks,
    xlabel style={font=\footnotesize},
    ylabel near ticks,
    ylabel style={font=\tiny},
    xmajorgrids,
    major x grid style={dotted},
    ymajorgrids,
    major y grid style={dotted},
    tick label style={font=\tiny}
  },
  mps scientific x/.style={
    x tick label style={
      /pgf/number format/sci,
      font=\tiny
    }
  },
  mps scientific y/.style={
    y tick label style={
      /pgf/number format/sci,
      font=\tiny
    }
  },
  mps fixed x/.style={
    x tick label style={
      /pgf/number format/.cd,
      fixed,
      fixed zerofill,
      precision=6,
      /tikz/.cd,
      font=\tiny
    }
  },
  mps fixed y/.style={
    y tick label style={
      /pgf/number format/.cd,
      fixed,
      fixed zerofill,
      precision=6,
      /tikz/.cd,
      font=\tiny
    }
  },
  /pgfplots/myylabel absolute/.style={%
      /pgfplots/every axis y label/.style={at={(0,0.5)},xshift=#1,rotate=90,align=center},
      /pgfplots/every y tick scale label/.style={
        at={(0,1)},above right,inner sep=0pt,yshift=0.3em
      }
   }
}
\usepackage{xcolor}
\definecolor{espblack}{RGB}{0,0,0}
\definecolor{espwhite}{RGB}{255,255,255}
\definecolor{espgray}{RGB}{206,206,206}
\definecolor{esplightgray}{RGB}{224,224,224}
\definecolor{espdarkgray}{RGB}{168,168,168}
\definecolor{espsomewhatdarkgray}{RGB}{130,130,130}
\definecolor{espverydarkgray}{RGB}{100,100,100}
\definecolor{espblue}{RGB}{11,93,174}
\definecolor{esplightblue}{RGB}{59,175,236}
\definecolor{espdarkblue}{RGB}{6,26,64}
\definecolor{espred}{RGB}{206,62,21}
\definecolor{esplightred}{RGB}{206,62,21}
\definecolor{espdarkred}{RGB}{61,19,8}
\definecolor{espyellow}{RGB}{232,163,26}
\definecolor{espgreen}{RGB}{100,161,27}
\definecolor{esplightgreen}{RGB}{149,198,35}
\definecolor{espdarkgreen}{RGB}{49,92,43}
\definecolor{esppurple}{RGB}{106,20,125}
\definecolor{esplightpurple}{RGB}{197,137,232}
\definecolor{espdarkpurple}{RGB}{50,14,59}

\usepackage{amsmath}
\usepackage{mathrsfs}
\usepackage{mathtools}
\usepackage{amsfonts}

\usepackage{subcaption}

\usepackage[capitalise]{cleveref}

\DeclareRobustCommand{\abbrevcrefs}{%
\crefname{algorithm}{Alg.}{Algs.}%
}

\DeclareRobustCommand{\cshref}[1]{{\abbrevcrefs\cref{#1}}}

\usepackage{siunitx}

\usepackage{multirow}

\usepackage{arydshln}

\usepackage{stackengine}

\usepackage[export]{adjustbox}

\newcommand{\setInsert}[0]{\xleftarrow{\scriptscriptstyle +}}

\title{ \bf
Efficient Path Planning In Manipulation Planning Problems by Actively Reusing Validation Effort
}

\author{Valentin N. Hartmann$^{1,2}$, Joaquim Ortiz-Haro$^{2}$, Marc Toussaint$^{2}$
\thanks{This research has been supported by the Deutsche Forschungsgemeinschaft (DFG, German Research Foundation) under Germany's Excellence Strategy -- EXC 2120/1–390831618, and by the German-Israeli Foundation for Scientific Research (GIF) grant I-1491-407.6/2019. Joaquim Ortiz-Haro thanks the International Max-Planck Research School for Intelligent
Systems for the support.
}%
\thanks{$^{1}$VISUS, University of Stuttgart, Germany
        {\tt\footnotesize \{firstname\}.\{lastname\}@ipvs.uni-stuttgart.de}
        }%
\thanks{$^{2}$Learning and Intelligent Systems Group, TU Berlin, Germany}%
}

\IEEEoverridecommandlockouts

\begin{document}
\maketitle

\begin{abstract}
The path planning problems arising in manipulation planning and in task and motion planning settings are typically repetitive: the same manipulator moves in a space that only changes slightly.
Despite this potential for reuse of information, few planners fully exploit the available information. %
To better enable this reuse, we decompose the collision checking into reusable, and non-reusable parts.
We then treat the sequences of path planning problems in manipulation planning as a multiquery path planning problem. %
This allows the usage of planners that \textit{actively} minimize planning effort over multiple queries, and by doing so, actively reuse previous knowledge. %

We implement this approach in EIRM* and effort ordered LazyPRM*, and benchmark it on multiple simulated robotic examples.
Further, we show that the approach of decomposing collision checks additionally enables the reuse of the gained knowledge over multiple different instances of the same problem, i.e., in a multiquery manipulation planning scenario.

The planners using the decomposed collision checking outperform the other planners in initial solution time by up to a factor of two while providing a similar solution quality.

\end{abstract}
\section{Introduction}

Manipulation planning requires planning both high level decisions on which actions to take (e.g., `pick obj$_1$ - place obj$_1$'), and low level planning, including planning of, e.g., grasping positions, and motion planning between start and goal configurations.
In this work, we focus on the motion planning part of the manipulation planning problem.
When solving the path planning problem of such manipulation planning problems, it is important that the methods are fast and reliable.
In path planning, checking if a proposed path is collision-free is usually the most time consuming part \cite{hauser2015lazy,sanchez2003single}, particularly when finding the initial solution \cite{Kleinbort2020}.

Manipulation planning problems are typically solved sequentially.
That is, in the example above, the planner first computes a path for the `pick'-action and then subsequently for the `place'-action.
This means that there is potential for reuse of gained information from previously solved planning queries.
To minimize overall planning time, it is desirable to \textit{actively} reuse information that we gained in the current and during previous planning queries.
However, directly reusing information of (in)valid motions is not possible since the environment can change during planning (e.g. by moving an obstacle), and so can the effective collision shape of a robot (by picking up and transporting an object), and previously valid motions can become invalid, or vice versa.

In most planners used for manipulation planning, knowledge gained during previous planning queries is discarded \cite{hauser2010task} (i.e. after already having planned the movement for picking up an object, the planner starts from scratch when searching for a path to place an object somewhere), or only used to bias the planning process in new queries \cite{Kingston2021b}.
The planner effectively needs to re-discover and re-validate the same edges repeatedly, wasting computational effort.

\begin{figure}[t]
    \centering
    \includegraphics[width=0.9\linewidth]{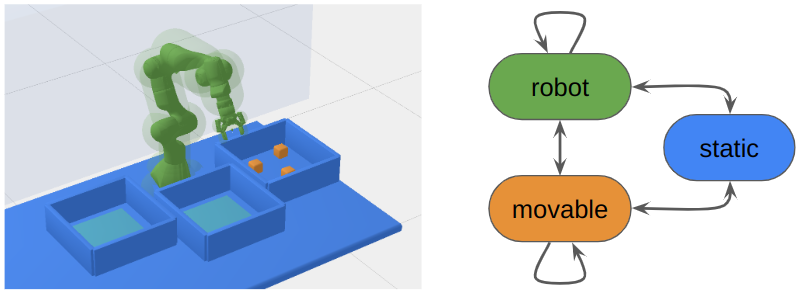}
    \caption{
    \textbf{Left}: Illustration of collision objects in decomposed collision checking.
    Everything in blue is \textit{static}, everything in orange is \textit{movable}, and everything in green is part of the \textit{robot}.
    \textbf{Right}: Arrows connect the groups that need to be collision checked against each other.
    }
    \label{fig:hierarchical_collision_checking}
\end{figure}

We propose an approach that enables the reuse of collision information throughout the complete manipulation planning problem.
We then treat the sequence of motion planning problems in a single manipulation planning problem as a multiquery motion planning problem.
By doing so, we are able to reuse collision information that was gained in previous planning queries.
We implement this approach in two planners that can actively reuse effort: Effort informed Roadmaps (EIRM*) \cite{hartmann_arxiv22} and a version of LazyPRM* with an effort ordered search.

By grouping all collision objects into \textit{i)} robot, \textit{ii)} movable objects, and \textit{iii)} static environment, a full collision check of a scene can be separated into five collision checks (\cref{fig:hierarchical_collision_checking}): robot-robot, robot-static environment, robot-movable objects, movable objects-movable objects, and static environment-movable objects.
Of these checks, the first two (robot-robot and robot-static environment) are reusable over all possible motion planning queries, as long as the robot and the static environment do not change.
This decomposition and reuse of information does not change completeness or optimality properties of the motion planning algorithms it is used in.

We demonstrate our approach on multiple simulated robotic problems.
We provide ablation studies, and compare planners using our approach to other state-of-the-art single-query planners that do not reuse information.

Finally, we also present our planner in a setting where similar manipulation planning problems have to be solved repeatedly, i.e., in a multiquery manipulation planning setting.
Our approach allows for reuse of gained information both among different actions in a single manipulation planning problem, and among multiple manipulation queries, and only requires minimal modifications in the planning algorithm.
\section{Related Work}

\subsection{Path planning in manipulation planning}
Manipulation planning is a special case of Task and Motion planning (TAMP) and is concerned with finding a sequence of actions, possibly their parametrizations, and corresponding paths to manipulate a set of objects into a desired final state.
During manipulation planning, the environment can be changed, e.g., by moving objects.
Examples for such problems often occur in construction \cite{hartmann2021long}, or assembly planning \cite{funk_learn2}.

Probabilistic tree-of-roadmaps (PTR) \cite{hauser2010task} randomly samples actions and then tries to find paths in the mode corresponding to the action.
The path planner used in PTR is a single-query planner that is instantiated newly for each action and attempt, and does not reuse any information.

A special case of a manipulation planning problem was presented in \cite{englert2021rss}, where the strong assumption is made that a transition does not change the free configuration space.
While that minimizes the number of required attempts that are needed to find a feasible solution, no information is reused between actions.

The knowledge from previous planning queries can be reused to bias new paths towards paths that are known valid from a previous planning query \cite{Kingston2021b, lai_2022, biomimetics7040210}.
While paths are biased towards known valid paths, previous (possibly reusable) collision checks are not reused.

Schmitt et al. \cite{schmitt2017optimal} present a manipulation planning approach that samples parametrizations of modes, and builds separate roadmaps for each parametrization.
The method then attempts to connect these separate roadmaps.
This approach can reuse previous plans, but it does so only in a very specific case, and not deliberately.

Task and Motion Informed Trees \cite{thomason_ral22} uses an asymmetric search to delay the computationally expensive operations, but does not explicitly reuse any gained information from the various solution attempts and does not reuse any information between the modes of the multi modal planning problem.

None of the presented approaches reuse validation effort in their path planning subroutine, as is possible with the approach we present.
The method we present can be used in the approaches discussed above to enable them to reuse previously invested effort as well.

\subsection{Multiquery path planning}
Multiquery path planning tries to efficienly solve multiple motion planning queries in the same (static) space.
Many multiquery planners are based on probabilistic roadmaps (PRM) \cite{kavraki1996probabilistic}, and variants thereof (e.g., (Lazy)PRM(*) \cite{karaman2011sampling,hauser2015lazy}).
These planners reuse previously invested planning effort if an edge is used in a path that was previously validated, but do not actively try to maximize reuse.
In practice this does not lead to much reuse in a planning query.

Lightning \cite{berenson2012robot} is a framework based on sparse roadmaps (SPARS) \cite{dobson2014sparse} that retrieves existing plans from a database and adapts them to the current setting.
As inserting and retrieving paths into and from SPARS can be computationally expensive, the overall planner might overall be slower than a planner that does not reuse information.

Experience graphs (E-Graphs) \cite{phillips2012graphs} actively reuse previously collision checked edges, but are computationally slow as an all-pairs shortest path algorithm needs to be ran repeatedly.

Reconfigurable Random Forest (RRF) \cite{li2002incremental} combines a graph search with tree-based methods to obtain a multiquery planner.
By growing new trees anchored at the start and goal respectively, and planning paths towards the existing other trees, a roadmap is built iteratively over the queries.
This approach does not give any optimality guarantees, and is not anytime.
Additionally, the roadmap that is constructed can grow indefinitely, which slows down subsequent planning queries.

EIRM* \cite{hartmann_arxiv22}, actively reuses previously invested computational effort by ordering its search for an initial solution by effort.
It uses a bidirectional search to first compute heuristics for a subsequent search of the underlying graph.
Compared to the other multiquery planners, EIRM* actively manages the graph growth that occurs over multiple queries.
EIRM* outperforms other multiquery planners on initial solution time by up to an order-of-magnitude.

None of these approaches can be applied directly to manipulation planning problems since they assume that the environment is static.
By decomposing the collision checking, and thus enabling reuse of validation effort, it is possible to reuse validation effort over multiple actions.

\section{Manipulation Planning as Multiquery Problem}

\subsection{Problem Statement \& Notation}
We consider a scene consisting of $N$ collision objects with poses $p\in\text{SE}(3)$.
We use $\mathcal{O}\subseteq \{1, ..., N\}$ to denote the movable objects and $\tau\in\mathcal{O}$ to refer to a specific object.
We then use $\theta_{\tau}\in\text{SE}(3)$ to denote the relative transformation of object $\tau$ to its parent frame, i.e. either to the gripper of the robot if an object is currently being manipulated, or to the static environment if it is at rest.

In manipulation planning, a sequence of actions is executed, such as `\texttt{pick(obj1)} - \texttt{place(obj1)}'\footnote{For a thorough discussion of the concept of an action in manipulation planning, refer to e.g. \cite{15-toussaint-IJCAI}.}.
We refer to $\Theta_i = \{\theta_\tau | \tau\in\mathcal{O}\}_i$ as the \textit{parametrization} of a scene during the $i$-th action.
We use $\mathcal{M}_i\subseteq\mathcal{O}$ to denote the set of objects that are manipulated (e.g. moved) during the $i$-th action of the sequence (in the previous pick and place example, $\mathcal{M}_1 = \{\}, \mathcal{M}_2=\{\texttt{obj1}\}$), and use $\theta_{\mathcal{M}_i}$ to denote the subset of the parametrization $\Theta_i$ that corresponds to the moving objects.

We use $\mathcal{C}\subseteq\mathbb{R}^n$ to denote the configuration space of the robot, and $q\in\mathcal{C}$ for a specific configuration of the robot\footnote{If multiple robots are present in the scene, we assume $\mathcal{C} = \mathcal{C}_1\times...\times\mathcal{C}_n$, i.e., the configuration-space is the stacked configuration-space of each robot.}.
The collision free configuration space during action $i$ is $\mathcal{C}_{\text{free}, i}(\Theta_i)$ and is dependent on the action $i$ that we are currently planning for and the parametrization of all the movable objects. 
The start state of the robot in action $i$ is $q_{i,\text{start}}$, and the goal state is $q_{i,\text{goal}}$.
While this approach easily extends to planning between sets of start and goal states, we only present the version with single states for ease of notation here.

\begin{figure}[t]
\centering
    \includegraphics[width=.45\linewidth]{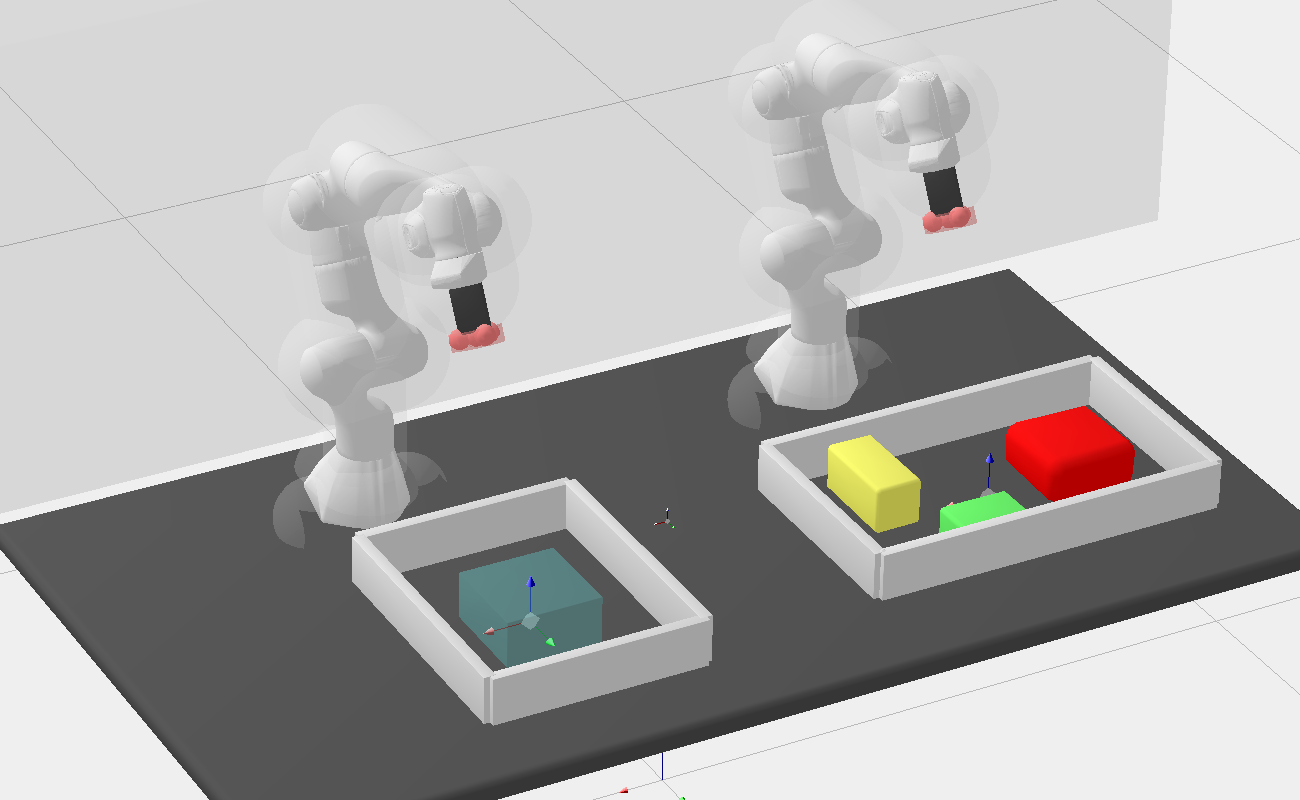}
    \includegraphics[width=.45\linewidth]{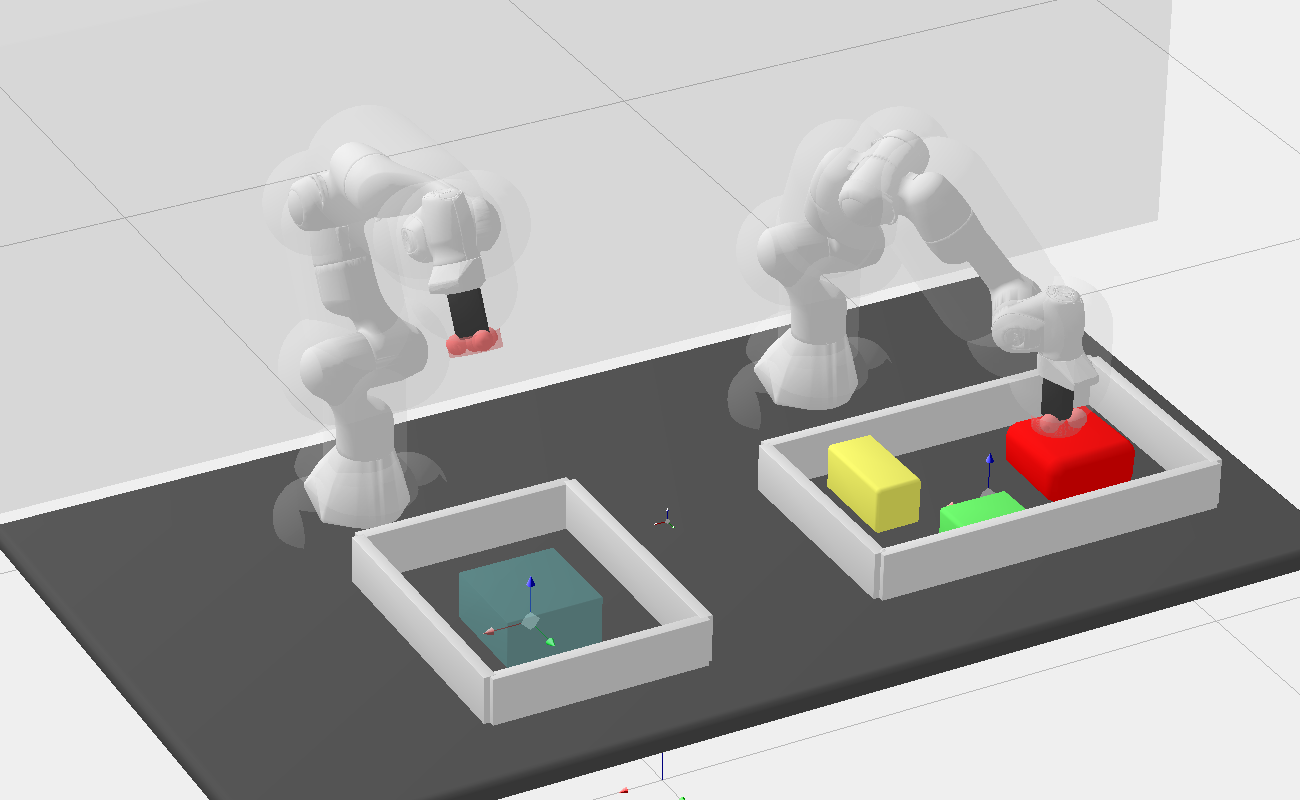}\\[1mm]
    \includegraphics[width=.45\linewidth]{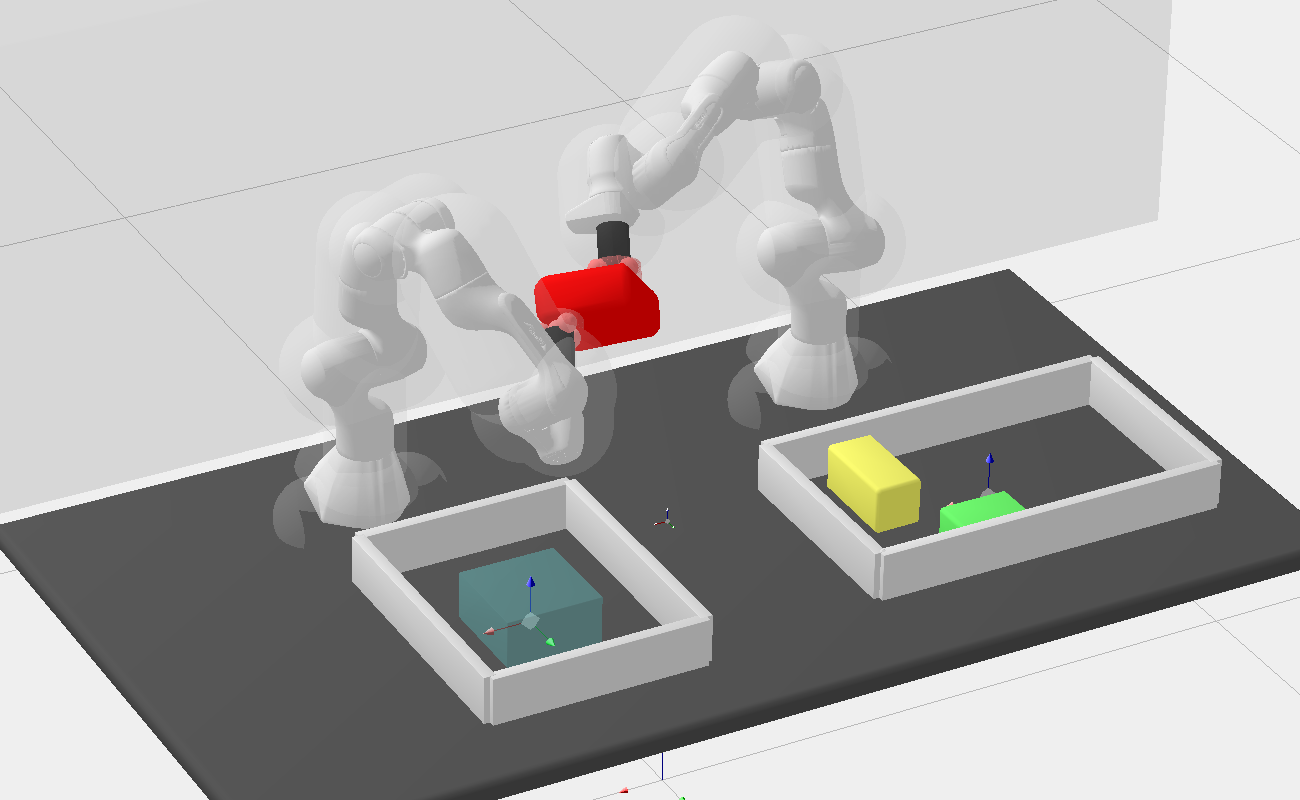}
    \includegraphics[width=.45\linewidth]{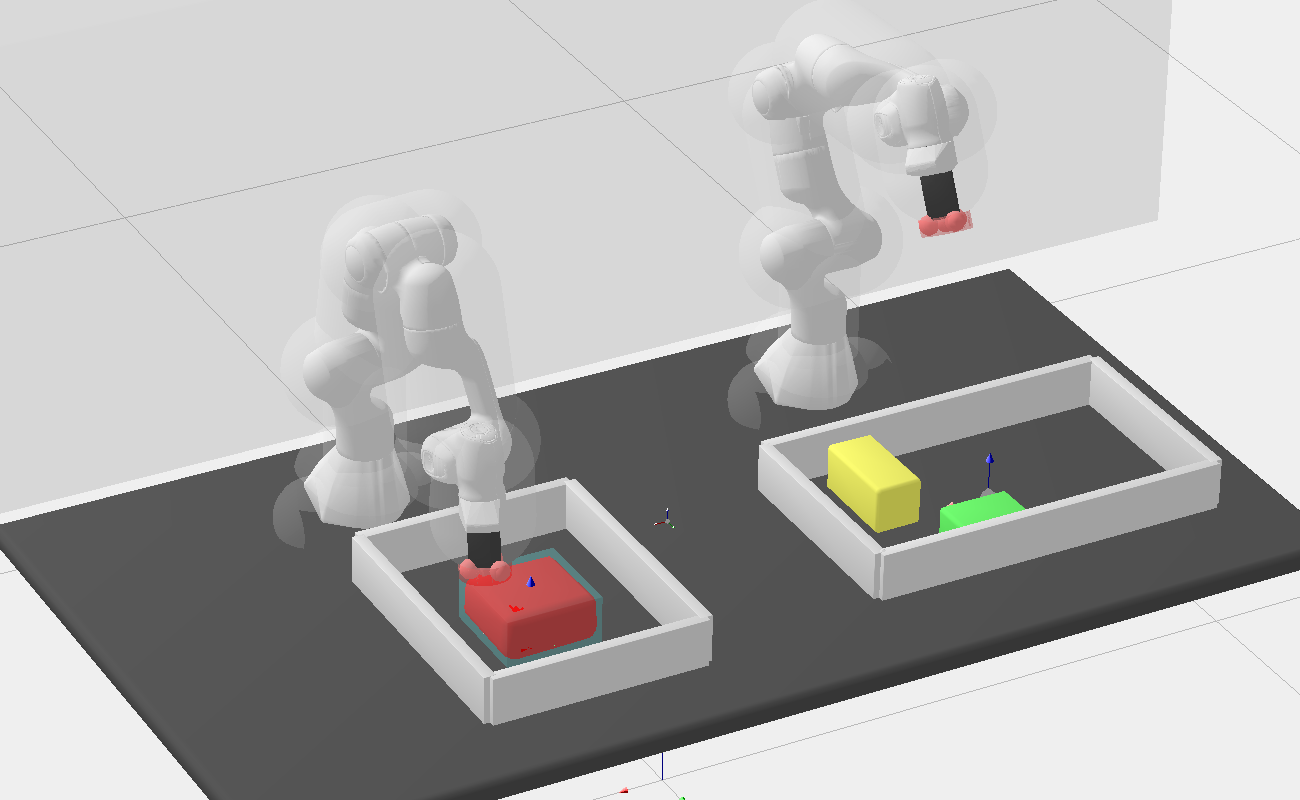}
    \caption{
    Start and goal configurations of the actions of a handover sequence.
    In this example, the manipulated objects in each action are $\mathcal{M}_1 = \{\}, \mathcal{M}_2 = \mathcal{M}_3 = \{\text{obj}_1\}$, and $\theta_1$ is the relative transformation of the red object to the table, and the relative transformation to the grippers of the robot arms, respectively.
    }
    \label{fig:manipulation_illustration}
\end{figure}

Given a tuple $\alpha_i = (q_{i,\text{start}}, q_{i,\text{goal}}, \mathcal{M}_i, \Theta_i)$ consisting of a start, a goal, the subset of objects that moves, and the parametrization of the movable objects, the goal is then to find a collision free path $\pi_i(s): [0, 1]\mapsto \mathcal{C}_{\text{free}, i}$ with $\pi_i(0) = q_{i,\text{start}}, \pi_i(1) = q_{i, \text{goal}}$.
Path planning in a manipulation planning problem then consists of a sequence $\mathcal{A}=\{\alpha_1, ..., \alpha_n\}$ of path planning problems, where the goal state coincides with the start state of the previous action, and the collision free environment at the end of the $i$-th action is equal to the beginning of the $(i+1)$-th action.
We show an example of the starts and goals of such a sequence in \cref{fig:manipulation_illustration}.
Computing the grasping positions in manipulation planning is an open research question in itself, and we assume that these positions are given.

We denote the planning effort between two states, $q$ and $q'$, using $e(q, q')$.
An example for the planning effort between two states is the computational effort it takes to check if the linear connection between two states is collision free.

We use $\setInsert$ for the addition of two sets, i.e., $A\setInsert B$ denotes $A\gets A\cup B$.

In some cases, it might be desirable to minimize a cost $c(\pi)$ over the path.
In this work, we use the path length as cost, but the presented approach works for any additive cost.
We denote the admissible estimate of the cost of a path going through state $q$ using $f(q)$.

\begin{algorithm}[t]
\caption{\texttt{IsValid}($q$, check$_\text{r/s}$, check$_\text{r/o}$, check$_\text{o/s}$)} 
\label{alg:main}
    $\mathcal{C}$.\texttt{SetConfiguration}($q$)\;
    coll$_\text{r/s}\gets$\texttt{false}; coll$_\text{r/o}\gets$\texttt{false}; coll$_\text{o/s}\gets$\texttt{false}\;
    \If{check$_\text{r/s}$\label{alg:main_check}}{
        coll$_\text{r/s}$ = $\mathcal{C}$.\texttt{CollideRobotStatic()} $\vee$ $\mathcal{C}$.\texttt{CollideRobotRobot()}\;
    }
    \If{check$_\text{r/o}$\label{alg:main_check_2}}{
        coll$_\text{r/o}$ = $\mathcal{C}$.\texttt{CollideRobotMovable()} $\vee$ $\mathcal{C}$.\texttt{CollideMovableMovable()}\;
    }
    \If{check$_\text{o/s}$\label{alg:main_check_3}}{
        coll$_\text{o/s}$ = $\mathcal{C}$.\texttt{CollideMovableStatic()}\;
    }
    \Return (coll$_\text{r/s}$, coll$_\text{r/o}$, coll$_\text{o/s}$)\;
\end{algorithm}

\begin{algorithm}[t]
\caption{\texttt{validate(path}, $\Theta_i, \mathcal{M}_i$\texttt{)} } 
\label{alg:validate}
    \For{$l\in$ \texttt{path.edges}}{
        check$_\text{r/s}$ $\gets l.\texttt{validity} ==$ \texttt{unknown}\label{alg:valIdate_check_re}\;
        check$_\text{o/s}$ $\gets l.\texttt{validity}[\theta_{\mathcal{M}_i}] ==$ \texttt{unknown}\label{alg:valIdate_check_oe}\;
        check$_\text{r/o}$ $\gets l.\texttt{validity}[\Theta_i] ==$ \texttt{unknown}\label{alg:valIdate_check_ro}\;
        \For{$q_\text{interp}\in$\texttt{interpolate}($l$, \texttt{resolution})\label{alg:validate_interpolate}}{
             (coll$_\text{r/s}$, coll$_\text{r/o}$, coll$_\text{o/s}$) $\gets$ \texttt{IsValid}($q_\text{interp}$, check$_\text{r/s}$, check$_\text{r/o}$, check$_\text{o/s}$)\label{alg:validate_check}\;
             \If{\textbf{any}(coll$_\text{r/s}$, coll$_\text{r/o}$, coll$_\text{o/s}$)}{
                \lIf{coll$_\text{r/s}$}{$l.\texttt{validity}\gets$ \texttt{invalid}\label{alg:valIdate_invalid}}
                \lIf{coll$_\text{o/s}$}{$l.\texttt{validity}[\theta_{\mathcal{M}_i}]\gets$ \texttt{invalid}\label{alg:valIdate_invalid_2}}
                \lIf{coll$_\text{r/o}$}{$l.\texttt{validity}[\Theta_i]\gets$ \texttt{invalid}\label{alg:valIdate_invalid_3}}
                \Return \texttt{false}\label{alg:valIdate_return_false}\;
             }
        }
        $l.\texttt{validity}\gets$ \texttt{valid}\label{alg:valIdate_valid}\;
        $l.\texttt{validity}[\theta_{\mathcal{M}_i}]\gets$ \texttt{valid}\label{alg:valIdate_valid_2}\;
        $l.\texttt{validity}[\Theta_i]\gets$ \texttt{valid}\label{alg:valIdate_valid_3}\;
    }
    \Return \texttt{true}\;
\end{algorithm}

\subsection{Decomposed collision checking}
The collision free configuration space in action $i$ of the manipulation planning problem can be written as  intersection of three collision free configuration spaces:
\begin{equation}\label{eq:coll_check}
    \mathcal{C}_{\text{free},i} = \mathcal{C}_\text{free}\cap \mathcal{C}_\text{free}(\theta_{\mathcal{M}_i})\cap \mathcal{C}_\text{free}(\Theta_i),
\end{equation}
where the first part is independent of the parametrization of the environment, the second part is only dependent on the objects that are moved in the current action, and the third part is dependent on all movable obstacles.

With the grouping introduced before (\cref{fig:hierarchical_collision_checking}), we obtain 5 collision checks that can be mapped to the decomposed configuration spaces (\cref{eq:coll_check}):
\begin{enumerate}
    \item $\mathcal{C}_\text{free}$: robot-robot, robot-static
    \item $\mathcal{C}_\text{free}(\theta_{\mathcal{M}_i})$: movable-static
    \item $\mathcal{C}_\text{free}(\Theta_i)$: movable-movable, robot-movable.
\end{enumerate}
As 1) is independent of both the action $i$ and the parametrization of the movable objects, the collision checking result can be reused in all planning queries.
Parts 2) and 3) can be reused if the parametrization $\Theta_i$, and the set of moving objects in the current action $\mathcal{M}_i$ is the same as in a previous collision check, respectively.

\subsection{Algorithms}
\Cref{alg:main,alg:validate} show how reusing knowledge of a collision check is implemented:
When collision checking an edge $l$, its \texttt{validity} for each of the parts of the decomposed collision check is either \texttt{unknown}, \texttt{valid}, or \texttt{invalid}.
If any of the validity components is \texttt{invalid}, the edge is invalid, and can thus not end up in a path found by the graph search of a planning algorithm.
If a part is known \texttt{valid}, the part does not have to be collision checked.
The validity of a part thus only has to be checked if it is \texttt{unknown} (\cshref{alg:validate}, \cref{alg:valIdate_check_re,alg:valIdate_check_oe,alg:valIdate_check_ro}, and \cshref{alg:main}, \cref{alg:main_check,alg:main_check_2,alg:main_check_3}).
An edge is then collision checked by interpolating the discrete points of the edge at a given \texttt{resolution} (\cshref{alg:validate}, \cref{alg:validate_interpolate}), and checking if the interpolated configuration is valid (\cshref{alg:validate}, \cref{alg:validate_check}).
Once an edge is completely collision checked and no collision was found, all its validity-components are annotated to be \texttt{valid} (\cshref{alg:validate}, \cref{alg:valIdate_valid,alg:valIdate_valid_2,alg:valIdate_valid_3}).
If an edge is in collision, the part that is in collision is marked as \texttt{invalid} (\cshref{alg:validate}, \cref{alg:valIdate_invalid,alg:valIdate_invalid_2,alg:valIdate_invalid_3}), and \texttt{validate} returns \texttt{false} (\cshref{alg:validate}, \cref{alg:valIdate_return_false}).

\begin{algorithm}[t]
\caption{Path planning in manipulation planning.} 
\label{alg:manip_planning}
    \texttt{paths}$\gets\emptyset$\;
    \For{$(q_{i,\text{start}}, q_{i,\text{goal}}, \mathcal{M}_i, \Theta_i)\in \mathcal{A}$}{
        $\mathcal{C}$.\texttt{UpdateParametrization}$(\mathcal{M}_i, \Theta_i)$\;
        \texttt{paths}$\setInsert$\texttt{plan}($q_{i,\text{start}}, q_{i,\text{goal}}, \mathcal{M}_i, \Theta_i$)\;
    }
    \Return \texttt{paths}\;
\end{algorithm}

\Cref{alg:manip_planning} shows how this reuse is implemented when solving a manipulation planning problem.
We demonstrate the decomposed collision checking in two path planning algorithms that are able to deliberately reuse effort: effort ordered LazyPRM* (\cshref{alg:helprm}) and EIRM* (\cshref{alg:eirm}).
In both planning algorithms, we assume that an edge-implict random geometric graph $X_\text{RGG}$ is used to store the samples that currently make up our approximation of the free space.
\paragraph{Effort Ordered Lazy PRM* (eo-LazyPRM*)}
We implement the PRM* based algorithm with sparse collision checking similar to \cite{strub_ijrr22}, that is, during the graph-search, the edges are checked sparsely.
Once a candidate path was found, it is then densely checked (using \cshref{alg:validate}).
We order the search initially by effort, with cost as a tie-breaker.
To achieve optimality, once an initial solution was found, the search is ordered by cost, with effort being the tiebreaker.

We additionally implement the graph size management approach that is implemented in EIRM*:
We use \textit{batch rewinding}: all sampled states are stored in a buffer $\mathcal{B}$, and a new batch of samples (of size $m$) is added from the buffer to the current graph once \texttt{refine\_approximation} (\cshref{alg:approximate}) is called.
At the beginning of a motion planning query, we start with an empty graph, and subsequently add the samples from the buffer $\mathcal{B}$ back to the graph.
For more details, refer to \cite{hartmann_arxiv22}.

\begin{algorithm}[t]
\caption{Effort ordered Lazy PRM*: \texttt{plan(}$q_{i, \text{start}},q_{i, \text{goal}}, \Theta_i, \mathcal{M}_i$\texttt{)}. Changes compared to LazyPRM* in {\color{orange}{orange}}.} 
\label{alg:helprm}
    $X_\text{RGG}\gets\{q_{i, \text{start}}, q_{i, \text{goal}}\}$; $c_\text{best}\gets\infty$; $\texttt{path}\gets\emptyset$; $i_\text{buffer} \gets 0$\;
    \While{\texttt{not stopped}}{
        $X_\text{RGG}\setInsert$\texttt{refine\_approximation(batch\_size, $c_\text{best}$)}\;
        \uIf{$c_\text{best}$ \textbf{is} $\infty$}{
            \color{orange}{$p\gets X_\text{RGG}$.\texttt{search\_effort\_ordered}()}\;
        }
        \Else{
            $p\gets X_\text{RGG}$.\texttt{search\_cost\_ordered}()\;
        }
        \If{{\color{orange}{\texttt{validate}($p, \Theta_i, \mathcal{M}_i$)}} \textbf{and} \texttt{cost(}$p$\texttt{)} $<c_\text{best}$}{
            $c_\text{best}\gets$ \texttt{cost(}$p$\texttt{)}\;
            $\texttt{path}\gets p$\;
        }
    }
    \Return \texttt{path}\;
\end{algorithm}

\paragraph{Effort Informed Roadmaps (EIRM*)}
EIRM* relies on an asymmetric bidirectional search, where a computationally inexpensive reverse search is used to compute problem and approximation specific heuristics.
These heuristics are then used to guide a forward search, which fully collision checks the edges.
The core changes compared to the implementation described in \cite{hartmann_arxiv22} are how collisions are checked, and how configurations are sampled (i.e. the usage of \cshref{alg:validate,alg:approximate} for validity checking, and refining the graph-approximation, respectively).

\begin{algorithm}[t]
\caption{EIRM*: \texttt{plan(}$q_{i, \text{start}},q_{i, \text{goal}}, \Theta_i, \mathcal{M}_i$\texttt{)}} 
\label{alg:eirm}
    $X_\text{RGG}\gets\{q_{i, \text{start}}, q_{i, \text{goal}}\}$; $c_\text{best}\gets\infty$; $\texttt{path}\gets\emptyset$; $i_\text{buffer} \gets 0$\;
    \While{\texttt{not stopped}}{
        \uIf{\texttt{best\_rev\_edge\_improves\_sol()}}{
            $X_\text{RGG}$.\texttt{reverse\_search()}\;
        }
        \uElseIf{\texttt{best\_fwd\_edge\_improves\_sol()}}{
            \texttt{path} $\gets X_\text{RGG}$.\texttt{forward\_search()}\;
        }
        \Else{
            $X_\text{RGG}\setInsert$\texttt{refine\_approximation(batch\_size, $c_\text{best}$)}\;
        }
    }
    \Return \texttt{path}\;
\end{algorithm}

\begin{algorithm}[t]
\caption{\texttt{refine\_approximation}($m, c_\text{best}$). Changes compared to original \cite{hartmann_arxiv22} in \color{orange}{orange}.} 
\label{alg:approximate}
    $\mathcal{B} \equiv \left(b_1, b_2, \ldots, b_n \right)$\; %
    $M \gets \emptyset$\;
    \While{$|M|\leq m$}
    {   
        \If {$i_\text{buffer} > |\mathcal{B}|$}
        {
            $q\gets$\texttt{sample\_uniform}()\;
            \color{orange}{\If{\texttt{IsValid}($q$, true, false, false)}{
                $\mathcal{B} \setInsert q$\;
            }}
        }
        \If{$f(b_{i_\text{buffer}})<c_\text{best}$ \textbf{and} \color{orange}{\texttt{IsValid}($q$, false, true, true)}\label{alg:rejection}}{
            $M \setInsert b_{i_\text{buffer}}$\label{alg:return_buffer}\;
        }
        $i_\text{buffer} \gets i_\text{buffer} + 1$\label{alg:increment}\;
    }
    \Return $M$\;
\end{algorithm}

\subsubsection{Effort estimate}
We use the validation effort as estimate for the planning effort $e$, i.e. the computational effort necessary to verify that an edge is collision free.
The planning effort $e$ between two states is then equal to the Euclidean distance between them, divided by the collision checking resolution.

Given that the collision check can be decomposed into three separate parts that can partially be reused, we can also decompose the validation effort in three parts.
We estimate the validation effort required for an edge by summing up the required estimated relative effort of each of the 3 collision checks, i.e.
\begin{equation}
    e(q, q') = e_{r/s}(q, q') + e_{o/s}(q, q'| \theta_{\mathcal{M}_i}) + e_{r/o}(q, q'| \Theta_i),
\end{equation}
where $e_{r/s}$ corresponds to the time required to check if an edge is in $\mathcal{C}_\text{free}$, $e_{o/s}$ corresponds to $\mathcal{C}_\text{free}(\theta_{\mathcal{M}_i})$, and $e_{r/o}$ corresponds to $\mathcal{C}_\text{free}(\Theta_i)$.
If the edge between $q, q'$ was previously checked and thus already annotated with its validity, that part of the sum is zero.

EIRM* requires an estimate for the \textit{a priori} effort-to-go $d$ (i.e. the effort to go from a target state to a goal state), which is used to order its reverse search.
Compared to \cite{hartmann_arxiv22}, where $d$ is 0, we use the more informative
\begin{equation}
    d(q) = 
        \max\{
            e(q, q_{i, \text{goal}}) - (e_\text{reusable} - e_\text{claimed}), 0 \}.
\end{equation}
We track the total reusable effort $e_\text{reusable}$ that was invested so far, and the effort $e_\text{claimed}$ that was claimed during the search so far (i.e. effort that was already reused).
The effort-to-go is then the difference between the effort $e(q, q_{i, \text{goal}})$ that it takes to validate the edge between the current state and the goal, and the remaining effort.
That is, we optimistically assume that all the remaining effort is part of the path to the goal.

\begin{figure*}[t]
\centering
    \begin{subfigure}[t]{.2\textwidth}
        \centering
        \includegraphics[width=.97\linewidth]{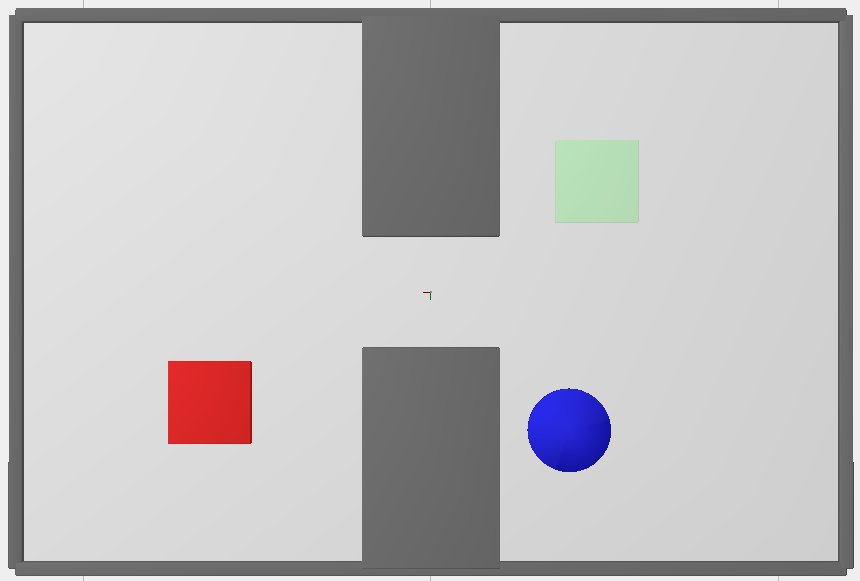}\\[1mm]
        \includegraphics[width=.97\linewidth]{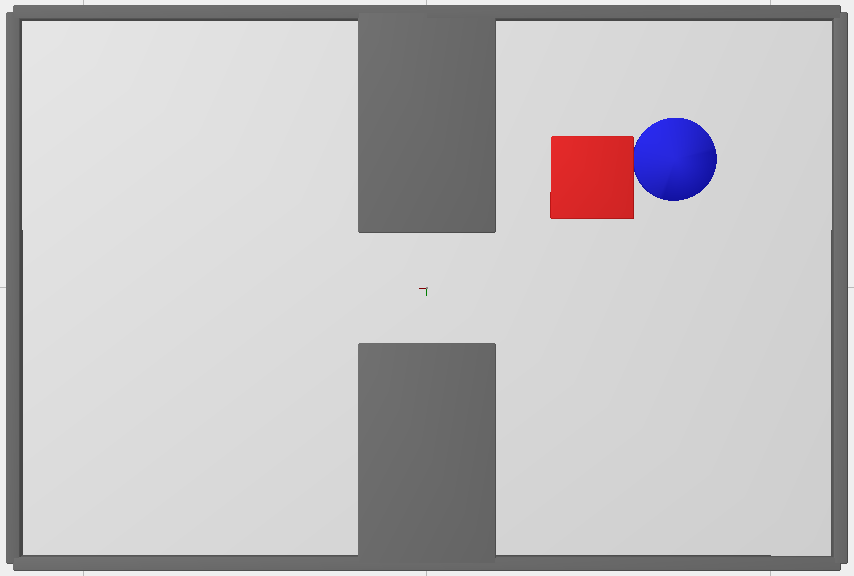}
        \caption{Wall Gap}
    \end{subfigure}%
    \begin{subfigure}[t]{.2\textwidth}
        \centering
        \includegraphics[width=.66\linewidth]{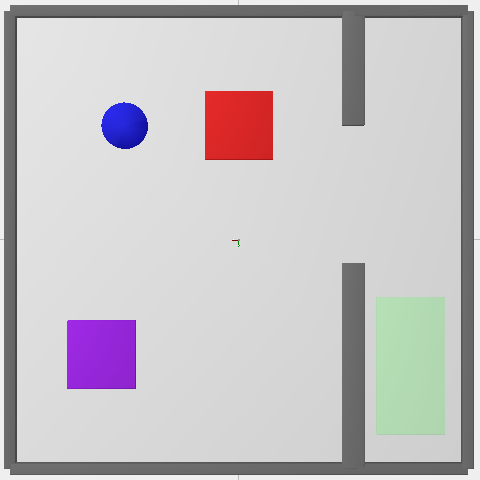}\\[1mm]
        \includegraphics[width=.66\linewidth]{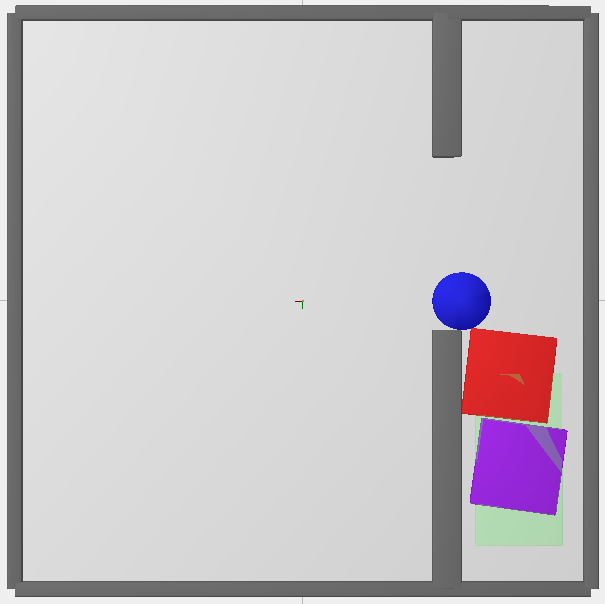}
        \caption{Multi Insertion}
    \end{subfigure}%
    \begin{subfigure}[t]{.2\textwidth}
        \centering
        \includegraphics[width=.97\linewidth]{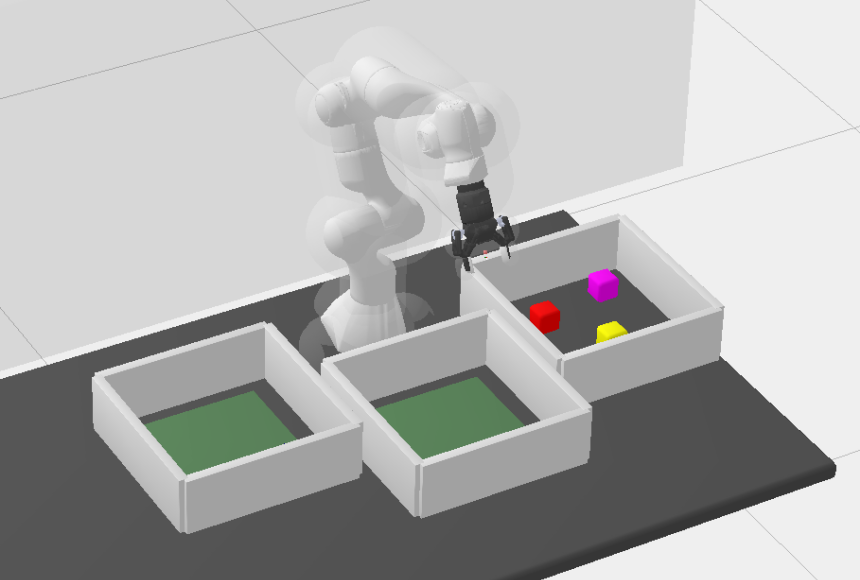}\\[1mm]
        \includegraphics[width=.97\linewidth]{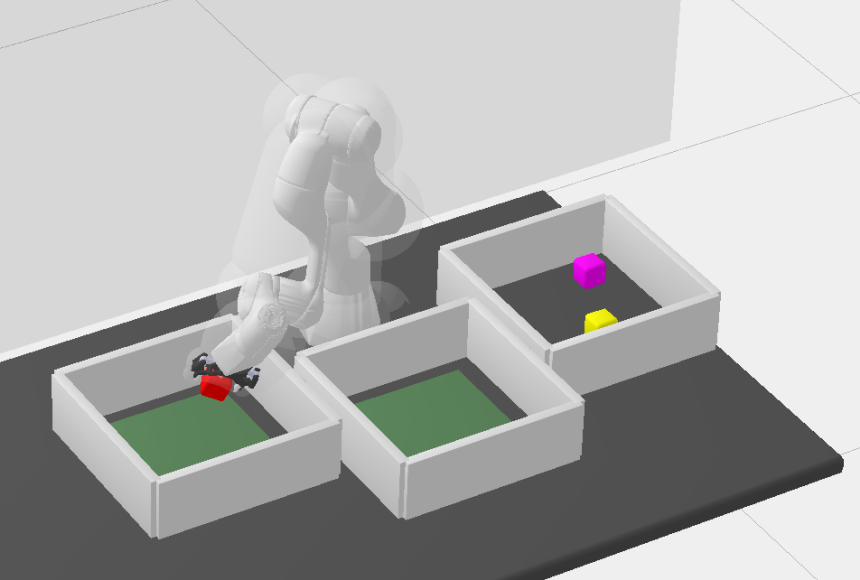}
        \caption{Bin Picking}
    \end{subfigure}%
    \begin{subfigure}[t]{.2\textwidth}
        \centering
        \includegraphics[width=.97\linewidth]{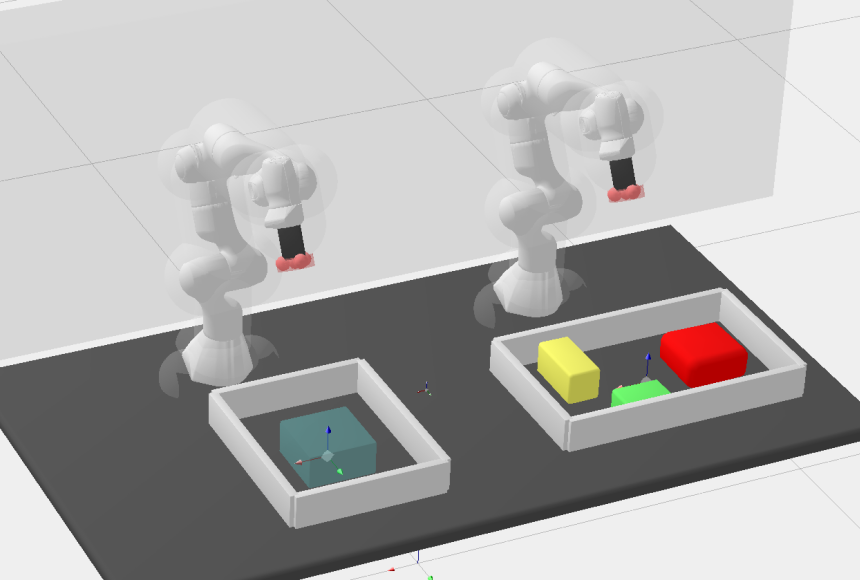}\\[1mm]
        \includegraphics[width=.97\linewidth]{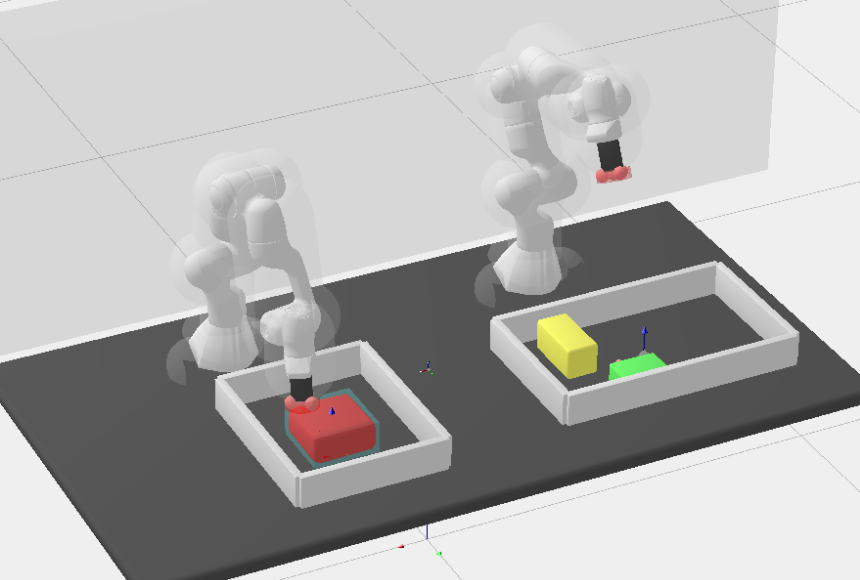}
        \caption{Handover}
    \end{subfigure}%
    \begin{subfigure}[t]{.2\textwidth}
        \centering
        \includegraphics[width=.97\linewidth]{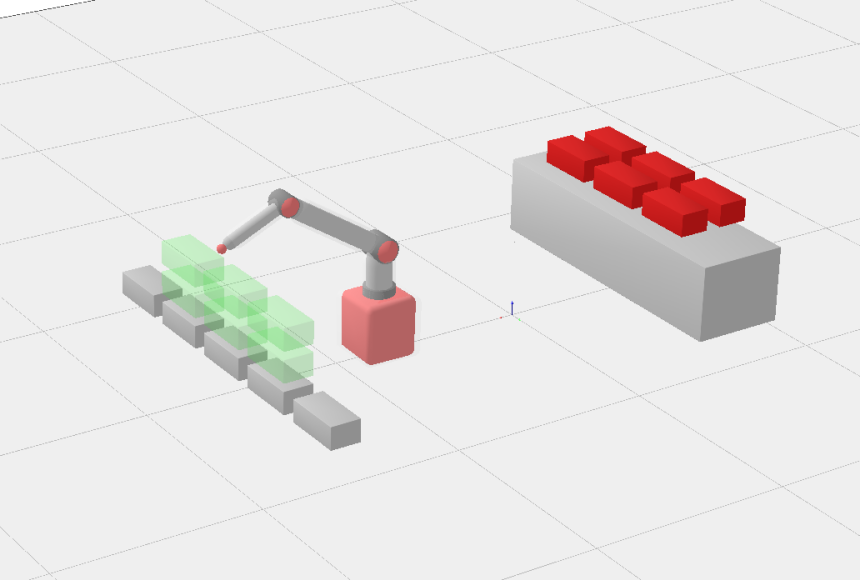}\\[1mm]
        \includegraphics[width=.97\linewidth]{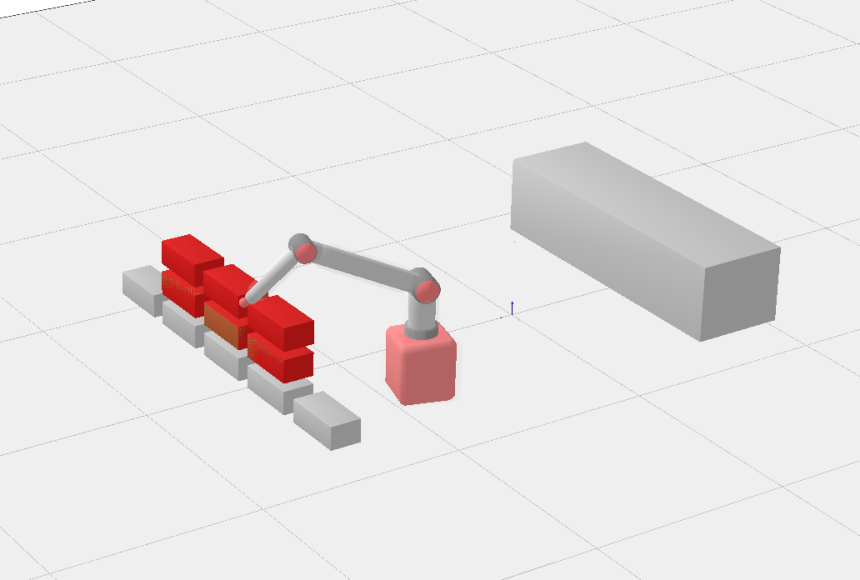}
        \caption{Brickwall}
    \end{subfigure}
    \caption{
    Images of the scenarios used in the benchmarks. 
    The start configuration is displayed at the top, and the goal configuration at the bottom. 
    }
    \label{fig:problems}
\end{figure*}

\section{Experiments}
We evaluate\footnote{All experiments were run on a laptop with an Intel i7-8565U CPU @ 1.80GHz processor with 16GB RAM. We make use of the benchmarking best-practices presented in \cite{gammell_empp22}.} our work on the scenarios that are shown in \cref{fig:problems}.
The action sequences are pick-and-place sequences, except for the \textit{Handover}-scenario, where the object is picked by the right arm, and handed over to the arm on the left, which then places the object at its goal location.

In the following scenarios, we assume that the grasp positions for the manipulation problems are given.
We compute the motions in the manipulation planning problems sequentially, i.e., we first solve the path planning problem for the first action, then for the second action, and so forth.

We use the Fast Collision Library (FCL) \cite{fcl} for our collision checks.
FCL natively offers the possibility to group collision-objects.
The relative time saved by decomposing the collision checks, and avoiding checking the robot-robot and robot-static env. interactions is between 20\% and 50\%.

\subsection{Algorithms \& Parameters}
We compare the planners using our approach against the OMPL \cite{sucan2012the-open-motion-planning-library} implementations of RRT-Connect and EIT* \cite{strub_ijrr22} as baselines that do not reuse any information between different queries.
Additionally, we show a version of LazyPRM* that uses the decomposed collision check, i.e. is able to reuse information, but orders its search by cost, and thus does not reuse information \textit{actively}.
To manage graph growth in this planner, we implement the same approach as in the effort ordered version.

We choose the collision checking resolution for the problems such that the false positive rate is below 1\textperthousand, i.e. one in thousand edges are though to be valid when they are actually invalid.
This results in roughly the collision checking resolutions of $10^{-4}$ for the abstract problems and $10^{-3}$ for the robotic problems.
For EIT*, EIRM*, and the LazyPRM*-variants, we use a batch size of 500 samples.

\subsection{Initial solution times and costs}
Each planner was run 100 times with different pseudorandom seeds.
The starts and goals were always the same during these 100 runs.
As we are interested in the time a planer takes to find the initial solution here, the planners were stopped as soon as a solution was found to the current path planning problem, and the next path planning problem in the manipulation planning sequence was started.

\begin{figure}[t]
\centering
    \begin{subfigure}[t]{.23\textwidth}
        \centering
        \includegraphics[width=.97\linewidth]{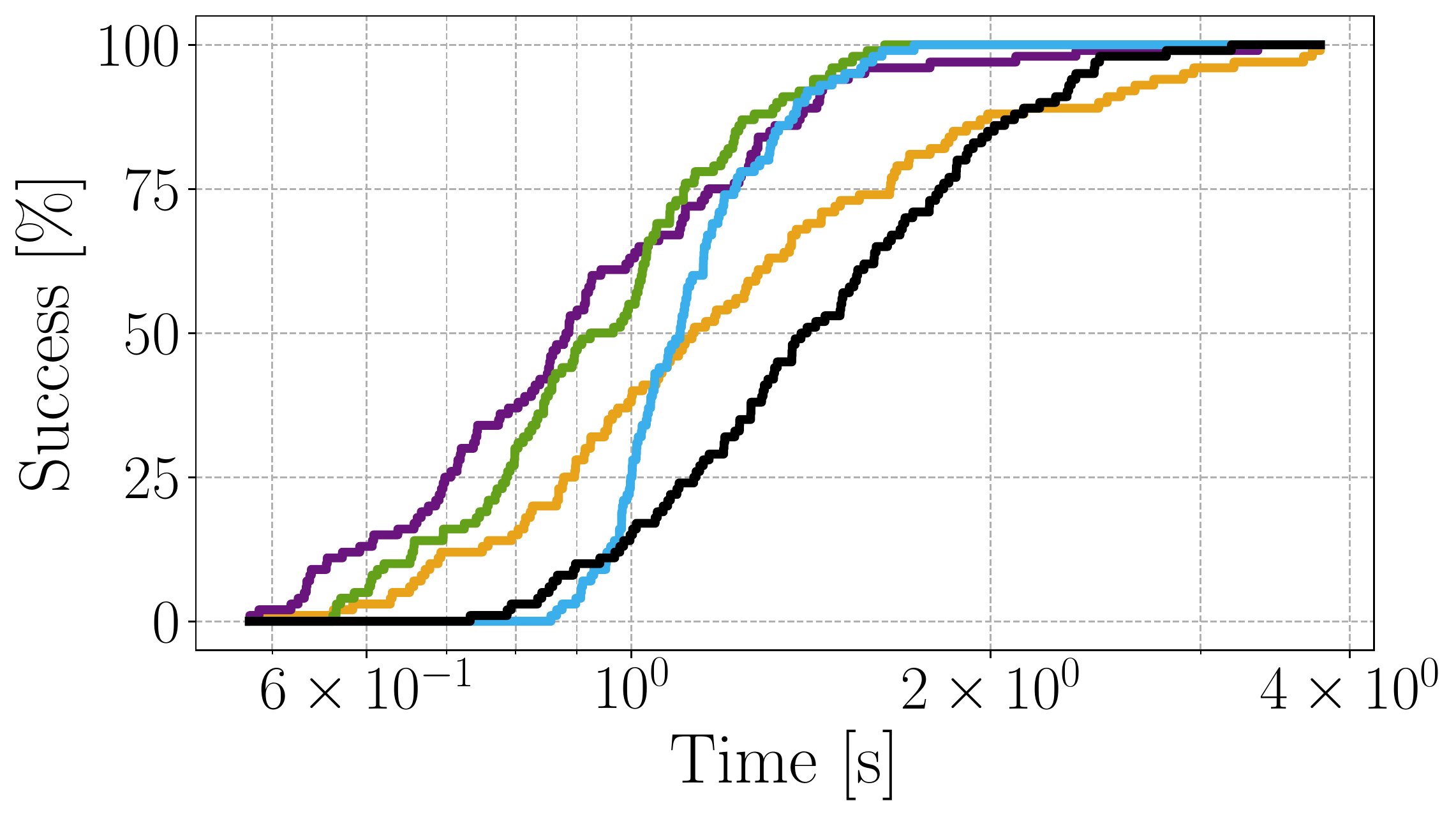}
        \caption{Bin Picking}
    \end{subfigure}%
    \begin{subfigure}[t]{.23\textwidth}
        \centering
        \includegraphics[width=.97\linewidth]{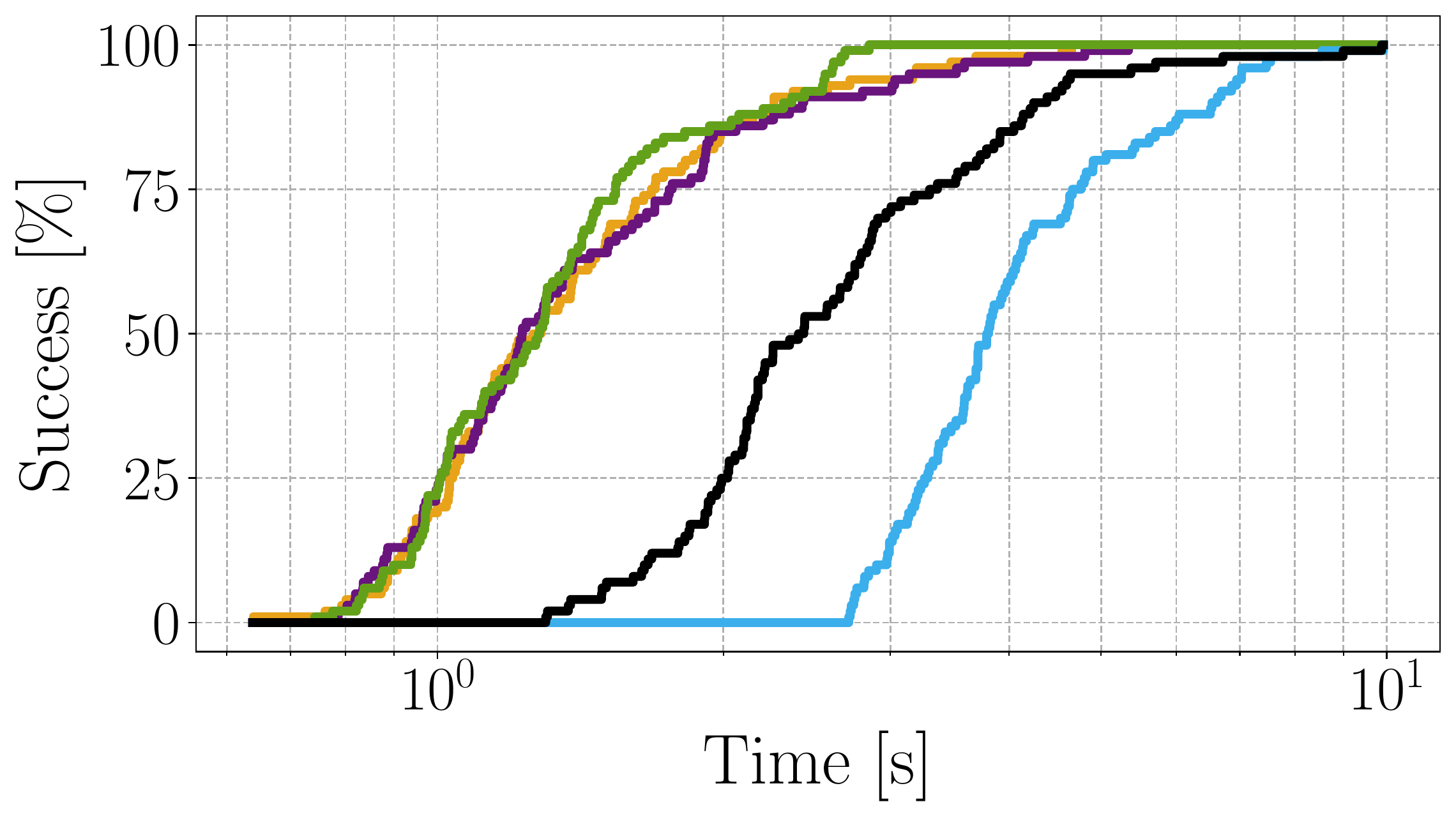}
        \caption{Handover}
    \end{subfigure}
    \\[0.5em]
    \begin{subfigure}[b]{1.0\linewidth}%
        \centering
        \begin{tikzpicture}
\begin{axis} [
  width=\textwidth,
  height=0.5\textwidth,
  unbounded coords=jump,
  xtick align=inside,
  ytick align=inside,
  anchor=north,
  hide axis,
  xmajorgrids,
  ymajorgrids,
  major grid style={densely dotted, black!20},
  xmin=0,
  xmax=10,
  ymin=0,
  ymax=10,
  xlabel style={font=\footnotesize},
  xticklabel style={font=\footnotesize},
  ylabel style={font=\footnotesize},
  yticklabel style={font=\footnotesize},
  legend style={anchor=south, legend cell align=left, legend columns=3, at={(axis cs:5, 6)}, font=\small}
]
\addlegendimage{espblack, line width = 1.0pt, mark size=1.0pt, mark=square*}
\addlegendentry{RRT-Connect}
\addlegendimage{esplightblue, line width = 1.0pt, mark size=1.0pt, mark=square*}
\addlegendentry{EIT*}
\addlegendimage{empty legend}
\addlegendentry{}
\addlegendimage{espyellow, line width = 1.0pt, mark size=1.0pt, mark=square*}
\addlegendentry{LazyPRM*}
\addlegendimage{esppurple, line width = 1.0pt, mark size=1.0pt, mark=square*}
\addlegendentry{eo-LazyPRM*}
\addlegendimage{espgreen, line width = 1.0pt, mark size=1.0pt, mark=square*}
\addlegendentry{EIRM*}

\end{axis}
\end{tikzpicture}
    \end{subfigure}
    \vspace{-5mm}
    \caption{
    The percentage of successful runs over time for each planner over the complete manipulation planning problem.
    }
    \label{fig:success}
\end{figure}

\Cref{tab:results} shows the median time to find the initial solution, and the corresponding median initial cost.
The success rate, i.e. the percentage of planners that found a solution for all the paths in the complete manipulation planning problem over time is shown in \cref{fig:success}.

The planners that reuse information outperform the planners that do not in initial solution time in all our tests.
The achieved cost of the initial solution of the planners that reuse information tends to be slightly lower than the one from the planners that do not reuse information.

\subsection{Cost evolution}
To analyze the cost convergence of the planners, we ran each planner 100 times for 10s on each planning query of the manipulation planning problem with different pseudorandom seeds.

\begin{figure}[t]
\centering
    \begin{subfigure}[t]{.23\textwidth}
        \centering
        \includegraphics[width=.97\linewidth, right]{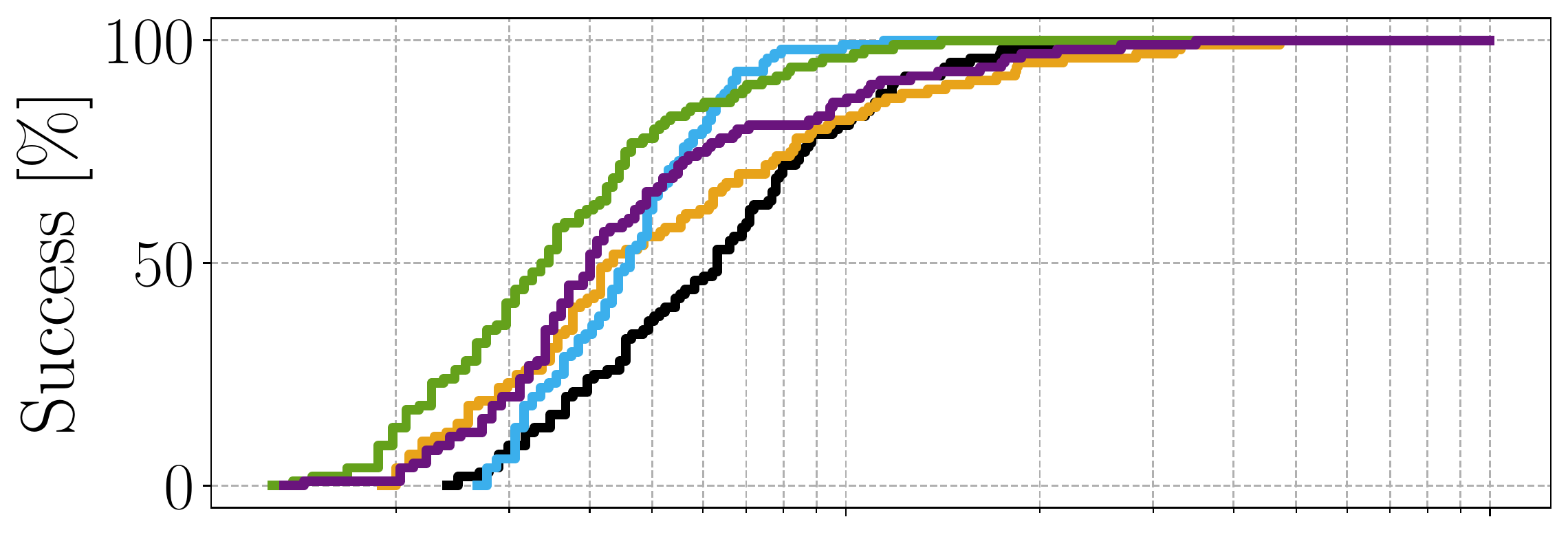}
        \includegraphics[width=.945\linewidth, right]{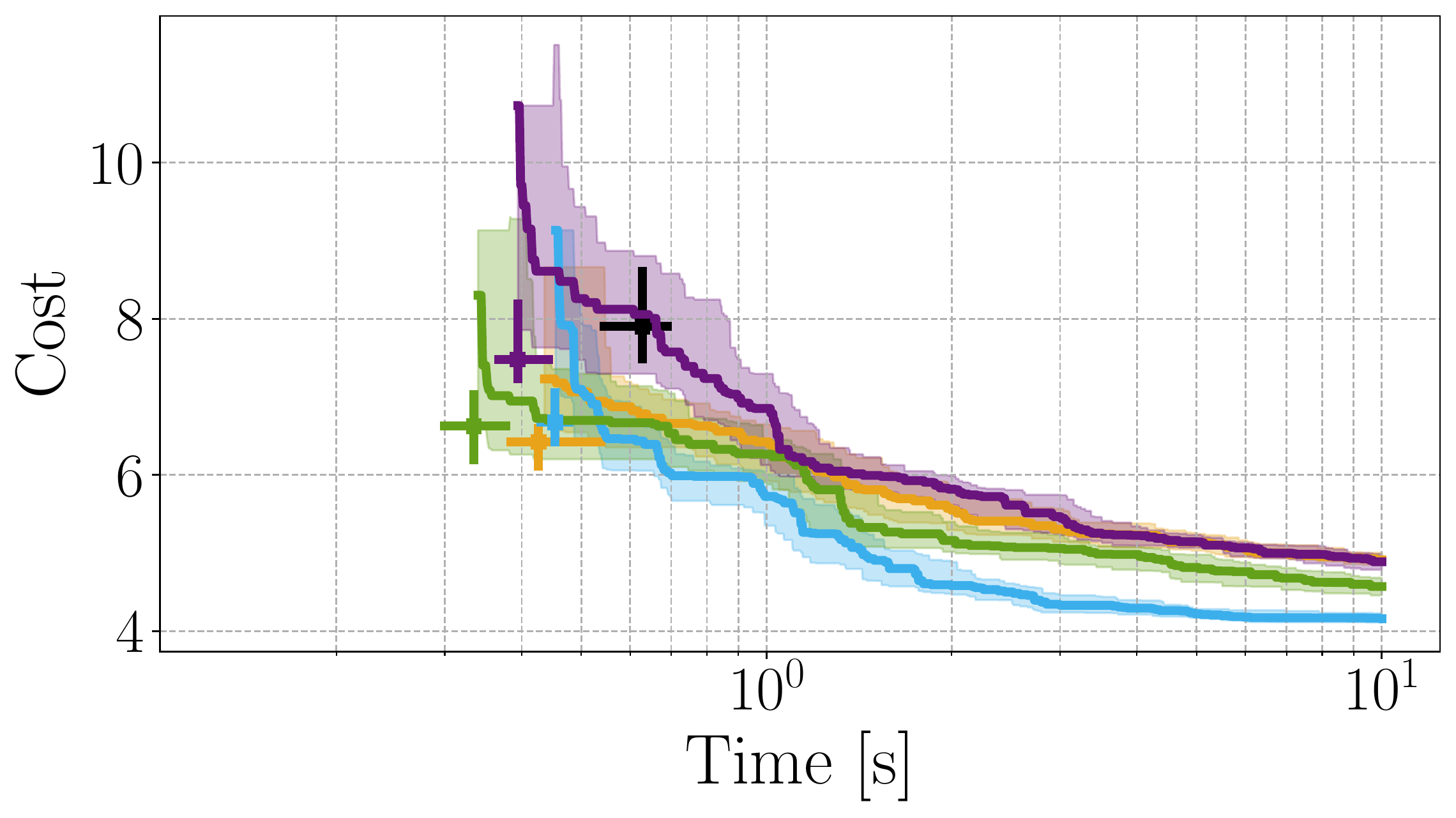}
        \caption{Bin Picking Action 1}
    \end{subfigure}%
    \begin{subfigure}[t]{.23\textwidth}
        \centering
        \includegraphics[width=.97\linewidth, right]{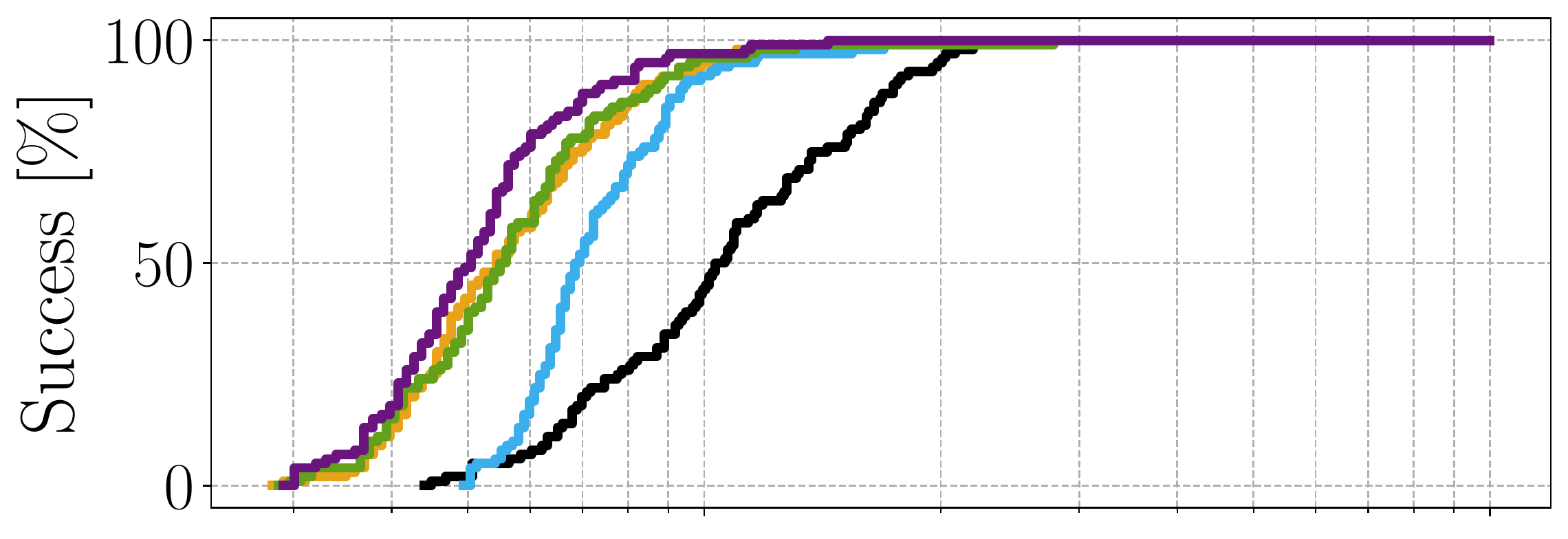}
        \includegraphics[width=.945\linewidth, right]{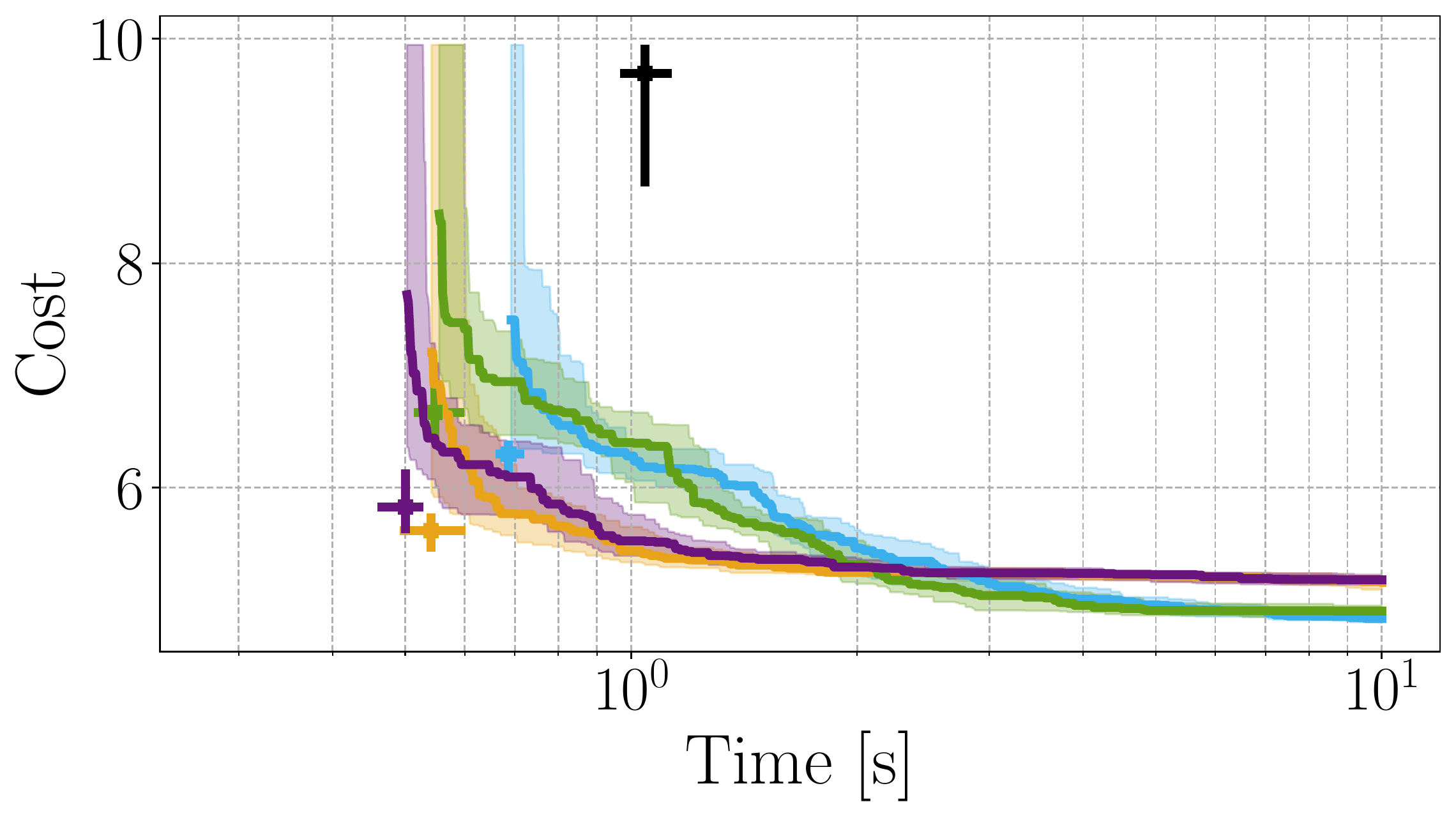}
        \caption{Bin Picking Action 2}
    \end{subfigure}
    \\[0.5em]
    \begin{subfigure}[b]{1.0\linewidth}%
        \centering
        \begin{tikzpicture}
\begin{axis} [
  width=\textwidth,
  height=0.5\textwidth,
  unbounded coords=jump,
  xtick align=inside,
  ytick align=inside,
  anchor=north,
  hide axis,
  xmajorgrids,
  ymajorgrids,
  major grid style={densely dotted, black!20},
  xmin=0,
  xmax=10,
  ymin=0,
  ymax=10,
  xlabel style={font=\footnotesize},
  xticklabel style={font=\footnotesize},
  ylabel style={font=\footnotesize},
  yticklabel style={font=\footnotesize},
  legend style={anchor=south, legend cell align=left, legend columns=3, at={(axis cs:5, 6)}, font=\small}
]
\addlegendimage{espblack, line width = 1.0pt, mark size=1.0pt, mark=square*}
\addlegendentry{RRT-Connect}
\addlegendimage{esplightblue, line width = 1.0pt, mark size=1.0pt, mark=square*}
\addlegendentry{EIT*}
\addlegendimage{empty legend}
\addlegendentry{}
\addlegendimage{espyellow, line width = 1.0pt, mark size=1.0pt, mark=square*}
\addlegendentry{LazyPRM*}
\addlegendimage{esppurple, line width = 1.0pt, mark size=1.0pt, mark=square*}
\addlegendentry{eo-LazyPRM*}
\addlegendimage{espgreen, line width = 1.0pt, mark size=1.0pt, mark=square*}
\addlegendentry{EIRM*}

\end{axis}
\end{tikzpicture}
    \end{subfigure}
    \vspace{-5mm}
    \caption{The cost evolution plots for the two modes of binpicking.
    The cost evolution plots (bottom) show the median cost at a given time as a thick line, with the nonparametric 95\% confidence interval as shaded area. 
    The squares show the median initial solution time for the query and the corresponding median initial cost. 
    The absence of a solution is treated as having an infinite cost.
    }
    \label{fig:cost}
\end{figure}

\Cref{fig:cost} shows the cost evolution of the solutions over time for each planner, and the corresponding success plot.
There is a noticeable speed-difference in the second action, where information can be reused by the multiquery planners.

The planners that reuse information converge to the optimal solution more slowly than EIT*.
Since the buffer needs to be filled with uniformly distributed samples to account for future queries, we can not directly sample the informed set. 
Thus, to sample the informed set, we use rejection sampling, which has a higher computational effort than directly sampling the informed set, once we get close to the optimal solution.

\begin{table*}[t]
\setstackgap{L}{6pt}
\setstackgap{S}{2pt}
\def\stackalignment{r}
\def\strutshortanchors{T}
\stackMath
\def\stacktype{S}

\renewrobustcmd{\bfseries}{\fontseries{b}\selectfont}
\renewrobustcmd{\boldmath}{}
\newrobustcmd{\B}{\bfseries}

    \centering
    \caption{Median time to initial solution and the corresponding initial cost from 100 runs.
    The sub and supscripts are the lower and upper nonparametric 95\% confidence bounds. 
    The best values for each problem are in \textbf{bold}.
    The planners above the dashed line are single-query planners, the planners below are multiquery planners using decomposed collision checking.
    }
    \sisetup{detect-weight=true,mode=text}
    {\renewcommand{\arraystretch}{1.2}
    \begin{tabular}{l|
        S[table-format=1.3]@{\hskip 0.6cm}
        S[table-format=2.2]@{\hskip 0.8cm}
        S[table-format=2.3]@{\hskip 0.6cm}
        S[table-format=2.2]@{\hskip 0.8cm}
        S[table-format=2.3]@{\hskip 0.6cm}
        S[table-format=2.2]@{\hskip 0.8cm}
        S[table-format=2.2]@{\hskip 0.6cm}
        S[table-format=3.2]@{\hskip 0.8cm}
        S[table-format=2.2]@{\hskip 0.6cm}
        S[table-format=2.2]
        }
        Scenario & \multicolumn{2}{c}{Wall Gap (2D)}  & \multicolumn{2}{c}{Multi Insertion (3D)} & \multicolumn{2}{c}{Binpicking (7D)} & \multicolumn{2}{c}{Brickwall (8D)} & \multicolumn{2}{c}{Handover (14D)} \\
        Number of Actions & \multicolumn{2}{c}{2}  & \multicolumn{2}{c}{4} & \multicolumn{2}{c}{2} & \multicolumn{2}{c}{12} & \multicolumn{2}{c}{3} \\
         & {$t_\text{init}$} & {$c_\text{init}$}& {$t_\text{init}$} & {$c_\text{init}$}& {$t_\text{init}$} & {$c_\text{init}$}& {$t_\text{init}$} & {$c_\text{init}$}& {$t_\text{init}$} & {$c_\text{init}$} \\
        \hline
RRT-Connect &  0.387\raisebox{0.1em}{\stackanchor{\scalebox{.6}{$+0.04$}}{\scalebox{.6}{$-0.04$}}} &  6.31 \raisebox{0.1em}{\stackanchor{\scalebox{.6}{$+0.34$}}{\scalebox{.6}{$-0.34$}}}&  0.728\raisebox{0.1em}{\stackanchor{\scalebox{.6}{$+0.08$}}{\scalebox{.6}{$-0.04$}}} &  28.26 \raisebox{0.1em}{\stackanchor{\scalebox{.6}{$+1.27$}}{\scalebox{.6}{$-1.21$}}}&  1.40\raisebox{0.1em}{\stackanchor{\scalebox{.6}{$+0.15$}}{\scalebox{.6}{$-0.10$}}} &  17.22 \raisebox{0.1em}{\stackanchor{\scalebox{.6}{$+0.90$}}{\scalebox{.6}{$-0.65$}}}&  5.48\raisebox{0.1em}{\stackanchor{\scalebox{.6}{$+0.37$}}{\scalebox{.6}{$-0.53$}}} &  141.37 \raisebox{0.1em}{\stackanchor{\scalebox{.6}{$+4.98$}}{\scalebox{.6}{$-4.27$}}}&  2.42\raisebox{0.1em}{\stackanchor{\scalebox{.6}{$+0.30$}}{\scalebox{.6}{$-0.24$}}} &  23.39 \raisebox{0.1em}{\stackanchor{\scalebox{.6}{$+0.43$}}{\scalebox{.6}{$-0.65$}}}\\
EIT* &  0.100\raisebox{0.1em}{\stackanchor{\scalebox{.6}{$+0.00$}}{\scalebox{.6}{$-0.00$}}} &  3.37 \raisebox{0.1em}{\stackanchor{\scalebox{.6}{$+0.01$}}{\scalebox{.6}{$-0.01$}}}&  0.127\raisebox{0.1em}{\stackanchor{\scalebox{.6}{$+0.01$}}{\scalebox{.6}{$-0.00$}}} &  10.71 \raisebox{0.1em}{\stackanchor{\scalebox{.6}{$+0.11$}}{\scalebox{.6}{$-0.40$}}}&  1.10\raisebox{0.1em}{\stackanchor{\scalebox{.6}{$+0.03$}}{\scalebox{.6}{$-0.05$}}} &  12.98 \raisebox{0.1em}{\stackanchor{\scalebox{.6}{$+0.49$}}{\scalebox{.6}{$-0.44$}}}&  1.73\raisebox{0.1em}{\stackanchor{\scalebox{.6}{$+0.06$}}{\scalebox{.6}{$-0.05$}}} &  58.59 \raisebox{0.1em}{\stackanchor{\scalebox{.6}{$+1.14$}}{\scalebox{.6}{$-2.47$}}}&  3.80\raisebox{0.1em}{\stackanchor{\scalebox{.6}{$+0.22$}}{\scalebox{.6}{$-0.18$}}} &  15.63 \raisebox{0.1em}{\stackanchor{\scalebox{.6}{$+0.97$}}{\scalebox{.6}{$-0.40$}}}\\
\hdashline
LazyPRM* &  0.206\raisebox{0.1em}{\stackanchor{\scalebox{.6}{$+0.01$}}{\scalebox{.6}{$-0.02$}}} & \bfseries 3.34 \raisebox{0.1em}{\stackanchor{\scalebox{.6}{$+0.01$}}{\scalebox{.6}{$-0.01$}}}&  0.178\raisebox{0.1em}{\stackanchor{\scalebox{.6}{$+0.01$}}{\scalebox{.6}{$-0.01$}}} & \bfseries 9.91 \raisebox{0.1em}{\stackanchor{\scalebox{.6}{$+0.13$}}{\scalebox{.6}{$-0.07$}}}&  1.13\raisebox{0.1em}{\stackanchor{\scalebox{.6}{$+0.15$}}{\scalebox{.6}{$-0.10$}}} & \bfseries 11.84 \raisebox{0.1em}{\stackanchor{\scalebox{.6}{$+0.48$}}{\scalebox{.6}{$-0.27$}}}&  2.49\raisebox{0.1em}{\stackanchor{\scalebox{.6}{$+0.36$}}{\scalebox{.6}{$-0.23$}}} & \bfseries 48.36 \raisebox{0.1em}{\stackanchor{\scalebox{.6}{$+1.50$}}{\scalebox{.6}{$-1.59$}}}&  1.27\raisebox{0.1em}{\stackanchor{\scalebox{.6}{$+0.11$}}{\scalebox{.6}{$-0.13$}}} & \bfseries 11.09 \raisebox{0.1em}{\stackanchor{\scalebox{.6}{$+0.26$}}{\scalebox{.6}{$-0.22$}}}\\
eo-LazyPRM* &  0.150\raisebox{0.1em}{\stackanchor{\scalebox{.6}{$+0.01$}}{\scalebox{.6}{$-0.01$}}} &  3.47 \raisebox{0.1em}{\stackanchor{\scalebox{.6}{$+0.03$}}{\scalebox{.6}{$-0.01$}}}&  0.163\raisebox{0.1em}{\stackanchor{\scalebox{.6}{$+0.01$}}{\scalebox{.6}{$-0.01$}}} &  11.94 \raisebox{0.1em}{\stackanchor{\scalebox{.6}{$+0.15$}}{\scalebox{.6}{$-0.35$}}}& \bfseries 0.884\raisebox{0.1em}{\stackanchor{\scalebox{.6}{$+0.04$}}{\scalebox{.6}{$-0.05$}}} &  13.47 \raisebox{0.1em}{\stackanchor{\scalebox{.6}{$+1.13$}}{\scalebox{.6}{$-0.57$}}}&  1.86\raisebox{0.1em}{\stackanchor{\scalebox{.6}{$+0.14$}}{\scalebox{.6}{$-0.13$}}} &  55.34 \raisebox{0.1em}{\stackanchor{\scalebox{.6}{$+2.92$}}{\scalebox{.6}{$-2.36$}}}& \bfseries 1.23\raisebox{0.1em}{\stackanchor{\scalebox{.6}{$+0.13$}}{\scalebox{.6}{$-0.06$}}} &  11.19 \raisebox{0.1em}{\stackanchor{\scalebox{.6}{$+0.94$}}{\scalebox{.6}{$-0.20$}}}\\
EIRM* & \bfseries 0.098\raisebox{0.1em}{\stackanchor{\scalebox{.6}{$+0.00$}}{\scalebox{.6}{$-0.00$}}} &  3.42 \raisebox{0.1em}{\stackanchor{\scalebox{.6}{$+0.02$}}{\scalebox{.6}{$-0.02$}}}& \bfseries 0.114\raisebox{0.1em}{\stackanchor{\scalebox{.6}{$+0.01$}}{\scalebox{.6}{$-0.00$}}} &  10.78 \raisebox{0.1em}{\stackanchor{\scalebox{.6}{$+0.33$}}{\scalebox{.6}{$-0.14$}}}&  0.945\raisebox{0.1em}{\stackanchor{\scalebox{.6}{$+0.07$}}{\scalebox{.6}{$-0.09$}}} &  13.93 \raisebox{0.1em}{\stackanchor{\scalebox{.6}{$+1.15$}}{\scalebox{.6}{$-0.46$}}}& \bfseries 1.55\raisebox{0.1em}{\stackanchor{\scalebox{.6}{$+0.04$}}{\scalebox{.6}{$-0.06$}}} &  51.25 \raisebox{0.1em}{\stackanchor{\scalebox{.6}{$+0.66$}}{\scalebox{.6}{$-1.85$}}}&  1.28\raisebox{0.1em}{\stackanchor{\scalebox{.6}{$+0.06$}}{\scalebox{.6}{$-0.14$}}} &  11.47 \raisebox{0.1em}{\stackanchor{\scalebox{.6}{$+0.64$}}{\scalebox{.6}{$-0.33$}}}\\
    \end{tabular}
    }
    \label{tab:results}
\end{table*}

\subsection{Multiquery manipulation planning}\label{ssec:mq}
We show an example where we solve multiple problems of the same \textit{family}, i.e., variations of the same problem where e.g. objects have different dimensions, or different initial and final positions.
Such problems often occur in e.g. in construction \cite{funk_learn2} or assembly planning \cite{hartmann2021long}.
Some additional care has to be exercised with reuse of collision information relating to the movable obstacles: if their shape changed, the collision information relating to them can not be reused.
Thus, only the information of the static environment and the robot were reused in this experiment.

\begin{figure}[t]
\centering
    \begin{subfigure}[t]{.23\textwidth}
        \centering
        \includegraphics[width=.97\linewidth]{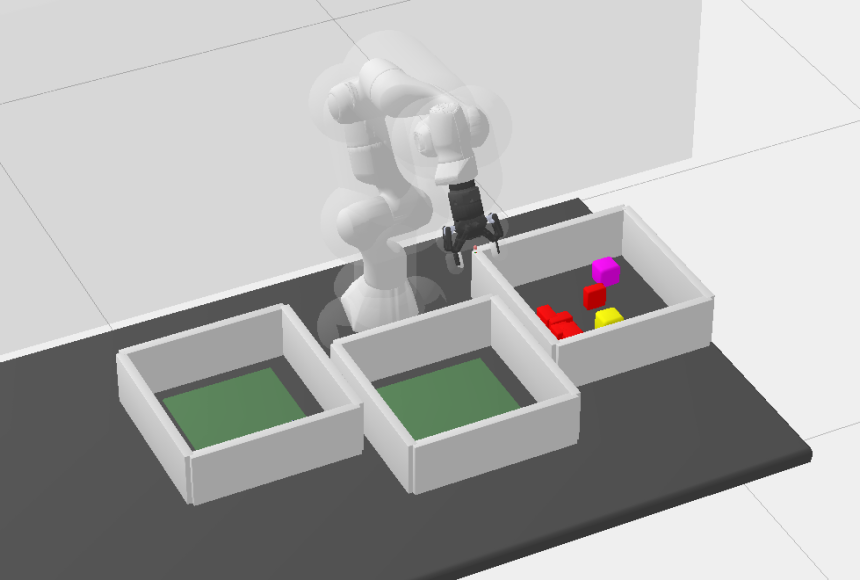}
        \caption{Bin Picking}
    \end{subfigure}%
    \begin{subfigure}[t]{.23\textwidth}
        \centering
        \includegraphics[width=.97\linewidth]{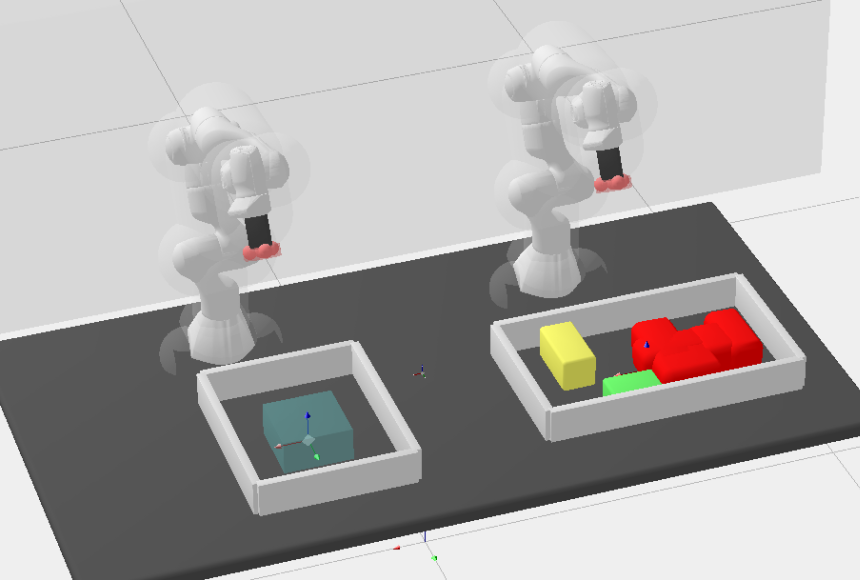}
        \caption{Handover}
    \end{subfigure}
        \caption{Illustration of some of the instances from a problem family with different object sizes and positions.}
    \label{fig:problem_families}
\end{figure}

We created 10 instances from the same problem family (i.e. scenarios where the initial and final goal, and the shape of the movable object changes), shown in \cref{fig:problem_families}.
Each planner was then run on the same 100 sequences made up by shuffling the 10 problem instances.

\begin{figure}[t]
\centering
    \begin{subfigure}[t]{.23\textwidth}
        \centering
        \includegraphics[width=.97\linewidth]{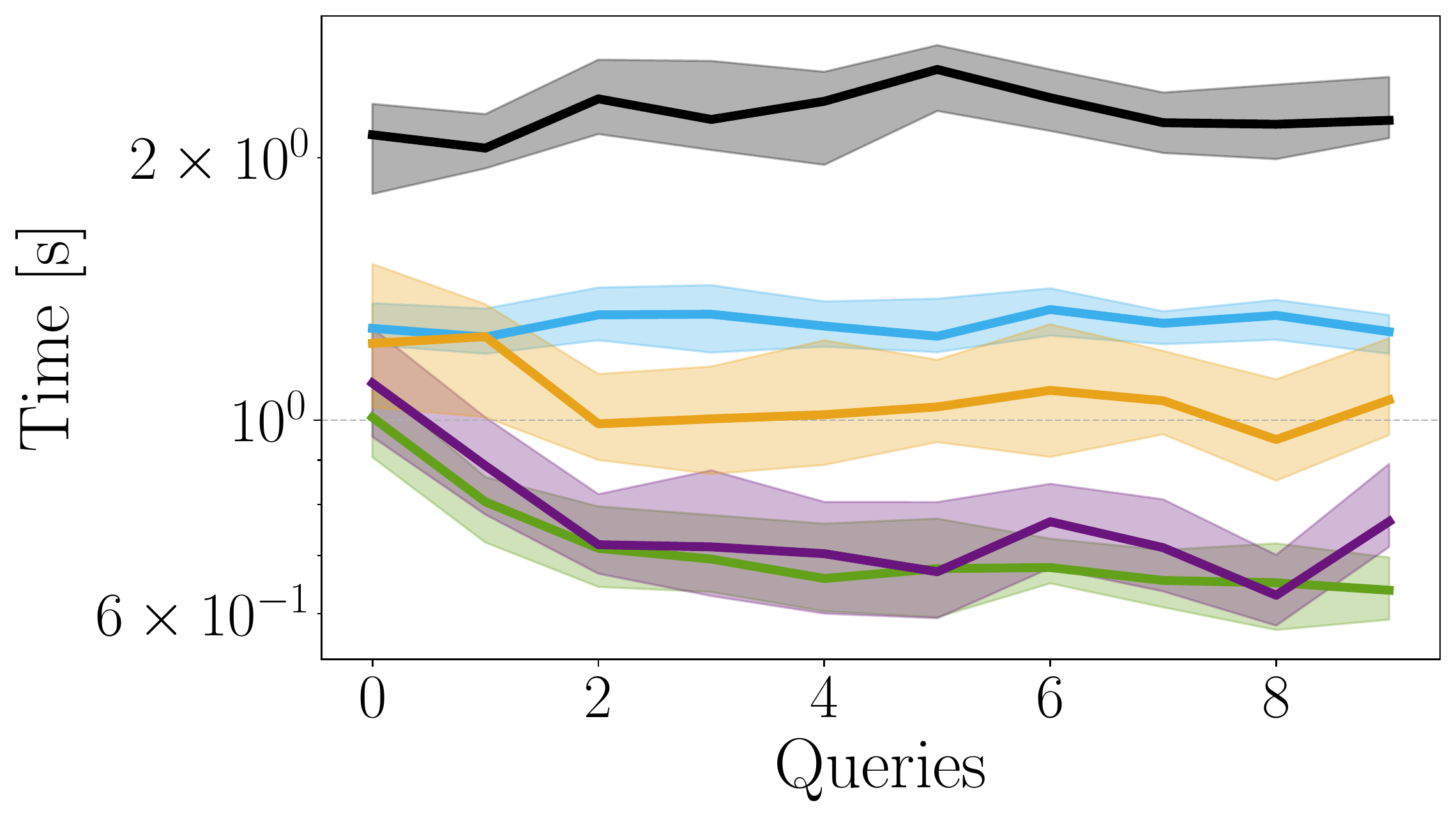}
        \caption{Bin Picking}
    \end{subfigure}%
    \begin{subfigure}[t]{.23\textwidth}
        \centering
        \includegraphics[width=.97\linewidth]{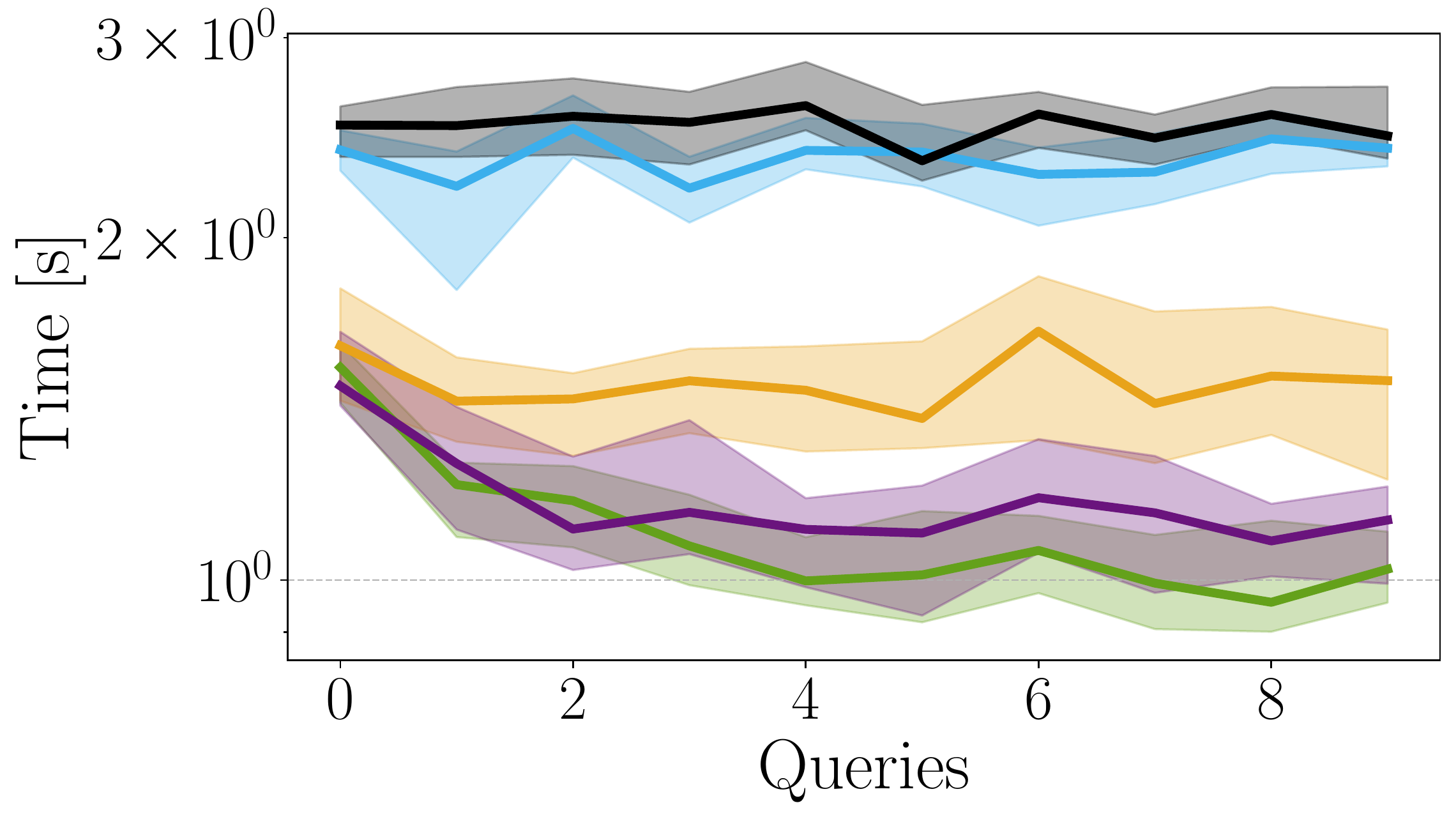}
        \caption{Handover}
    \end{subfigure}
    \\[0.5em]
    \begin{subfigure}[b]{1.0\linewidth}%
        \centering
        \begin{tikzpicture}
\begin{axis} [
  width=\textwidth,
  height=0.5\textwidth,
  unbounded coords=jump,
  xtick align=inside,
  ytick align=inside,
  anchor=north,
  hide axis,
  xmajorgrids,
  ymajorgrids,
  major grid style={densely dotted, black!20},
  xmin=0,
  xmax=10,
  ymin=0,
  ymax=10,
  xlabel style={font=\footnotesize},
  xticklabel style={font=\footnotesize},
  ylabel style={font=\footnotesize},
  yticklabel style={font=\footnotesize},
  legend style={anchor=south, legend cell align=left, legend columns=3, at={(axis cs:5, 6)}, font=\small}
]
\addlegendimage{espblack, line width = 1.0pt, mark size=1.0pt, mark=square*}
\addlegendentry{RRT-Connect}
\addlegendimage{esplightblue, line width = 1.0pt, mark size=1.0pt, mark=square*}
\addlegendentry{EIT*}
\addlegendimage{empty legend}
\addlegendentry{}
\addlegendimage{espyellow, line width = 1.0pt, mark size=1.0pt, mark=square*}
\addlegendentry{LazyPRM*}
\addlegendimage{esppurple, line width = 1.0pt, mark size=1.0pt, mark=square*}
\addlegendentry{eo-LazyPRM*}
\addlegendimage{espgreen, line width = 1.0pt, mark size=1.0pt, mark=square*}
\addlegendentry{EIRM*}

\end{axis}
\end{tikzpicture}
    \end{subfigure}
    \vspace{-5mm}
    \caption{
    Median motion planning times over multiple manipulation planning queries. 
    The thick line indicates the median planning time from 100 runs, and the shaded area shows the nonparametric 95\% confidence intervals.
    }
    \label{fig:multiquery_times}
\end{figure}

\Cref{fig:multiquery_times} shows the median initial solution time per query.
The planners using the decomposed collision checking outperform the single-query planners by approximately a factor of 2.5.
Of the multiquery planners, the ones that actively reuse previously invested effort outperform the one that does not.

\subsection{Ablations}
\subsubsection{Graph size management}
We compare the algorithms using effort ordered search to versions of them not using batch rewinding, i.e., we start with all valid samples that were used to find the previous solution.

\begin{figure}[t]
\centering
    \begin{subfigure}[t]{.23\textwidth}
        \centering
        \includegraphics[width=.97\linewidth]{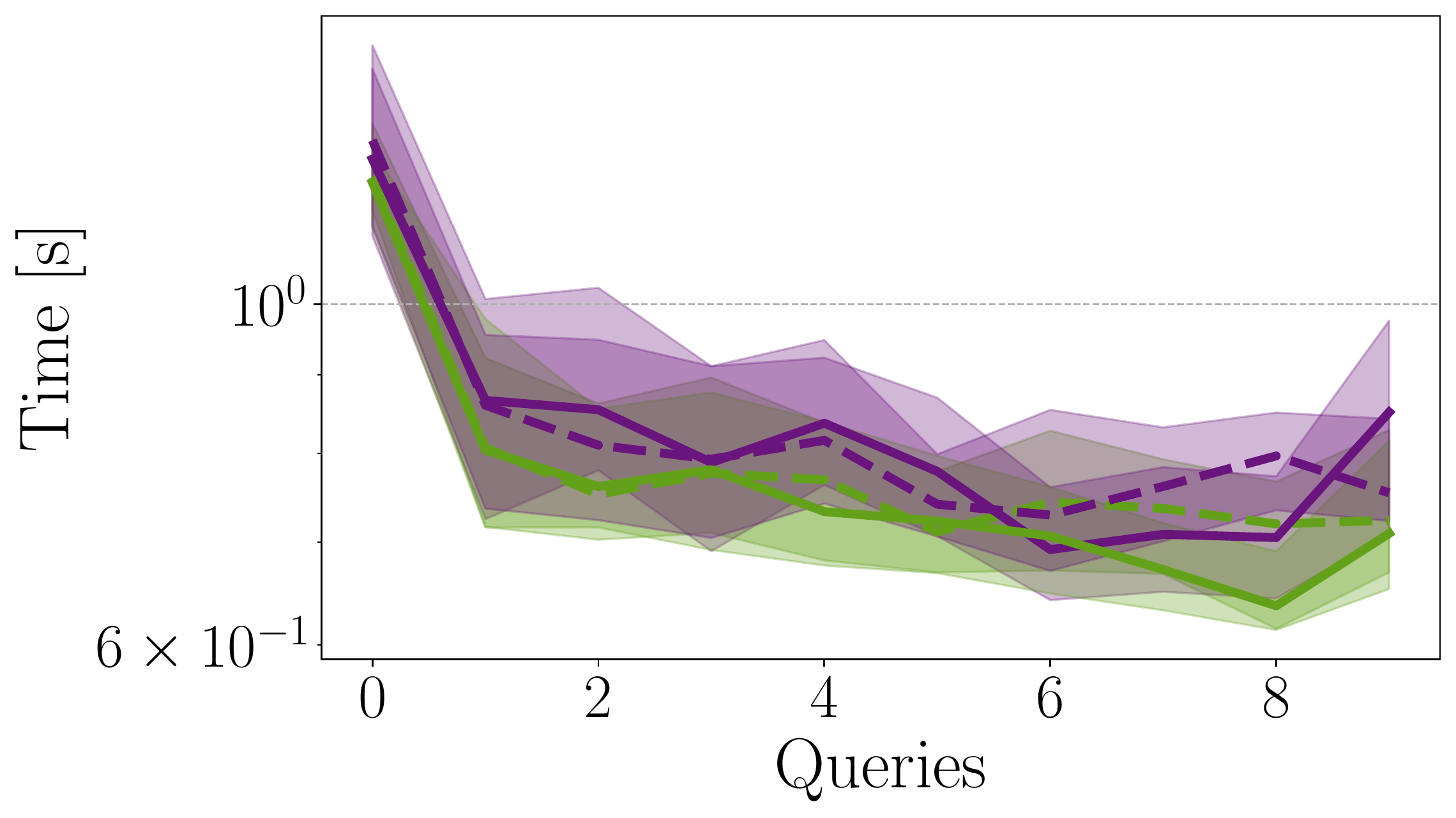}
        \caption{Bin Picking}
    \end{subfigure}%
    \begin{subfigure}[t]{.23\textwidth}
        \centering
        \includegraphics[width=.97\linewidth]{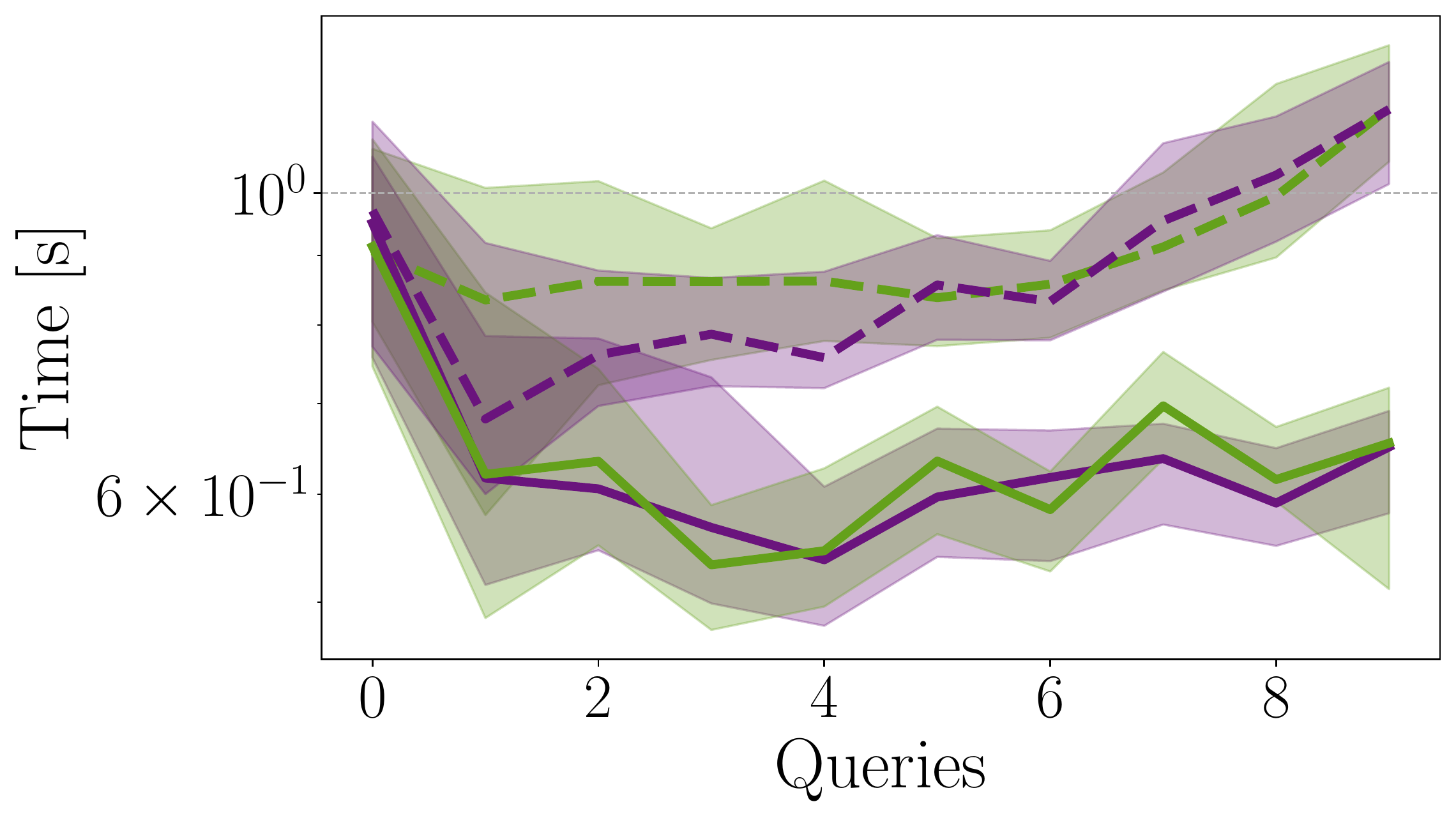}
        \caption{Handover}
    \end{subfigure}
    \\[0.5em]
    \begin{subfigure}[b]{1.0\linewidth}%
        \centering
        \begin{tikzpicture}
\begin{axis} [
  width=\textwidth,
  height=0.5\textwidth,
  unbounded coords=jump,
  xtick align=inside,
  ytick align=inside,
  anchor=north,
  hide axis,
  xmajorgrids,
  ymajorgrids,
  major grid style={densely dotted, black!20},
  xmin=0,
  xmax=10,
  ymin=0,
  ymax=10,
  xlabel style={font=\footnotesize},
  xticklabel style={font=\footnotesize},
  ylabel style={font=\footnotesize},
  yticklabel style={font=\footnotesize},
  legend style={anchor=south, legend cell align=left, legend columns=2, at={(axis cs:5, 6)}, font=\small}
]
\addlegendimage{esppurple, line width = 1.0pt, mark size=1.0pt, mark=square*}
\addlegendentry{eo-LazyPRM*}
\addlegendimage{espgreen, line width = 1.0pt, mark size=1.0pt, mark=square*}
\addlegendentry{EIRM*}

\addlegendimage{esppurple, dashed, line width = 1.0pt, mark size=1.0pt, mark=square*}
\addlegendentry{eo-LazyPRM* (unmanaged)}
\addlegendimage{espgreen, dashed, line width = 1.0pt, mark size=1.0pt, mark=square*}
\addlegendentry{EIRM* (unmanaged)}

\end{axis}
\end{tikzpicture}
    \end{subfigure}
    \vspace{-5mm}
    \caption{
    Comparison of planners that use \textit{batch rewinding}, and versions that do not on the same scenarios as in \cref{ssec:mq}.
    The thick line indicates the median planning time from 100 runs, and the shaded area shows the non-parameteric 95\% confidence intervals.
    }
    \label{fig:multiquery_ablation}
\end{figure}

\Cref{fig:multiquery_ablation} shows the median initial solution time for the same experiments as discussed in \cref{ssec:mq} for the algorithms without graph size management.

While in the scenario in on the left, the performance was compareable, in the handover scenario, the unmanaged version performs considerably worse.

Note that in the experiments here, these path planning runs were stopped once an initial solution is found.
In case the planner continues after the initial solution to optimize the path, the effect that can be seen here is stronger, and it can even be the case that no solution is found anymore in the given time if no active graph size management is done, since the density of the graph becomes too high (as discussed in \cite{hartmann_arxiv22}).

\subsubsection{Active vs. passive effort reuse}
We compare a planner that uses the decomposed collision checks but uses a cost ordered search, as is standard in LazyPRM*.
Thus, while the planner can reuse information it does not actively try to do so.
It is visible both in \cref{tab:results}, and \cref{fig:multiquery_times}, that the solution times that are obtained by the effort ordered search are smaller than the ones obtained by using the cost ordered search.
However, the cost ordered search leads to a smaller cost.

\subsubsection{Influence of collision checking resolution}\label{ssec:coll_ablation}
\begin{figure}[t]
    \begin{subfigure}[t]{.23\textwidth}
        \centering
        \includegraphics[width=.97\linewidth]{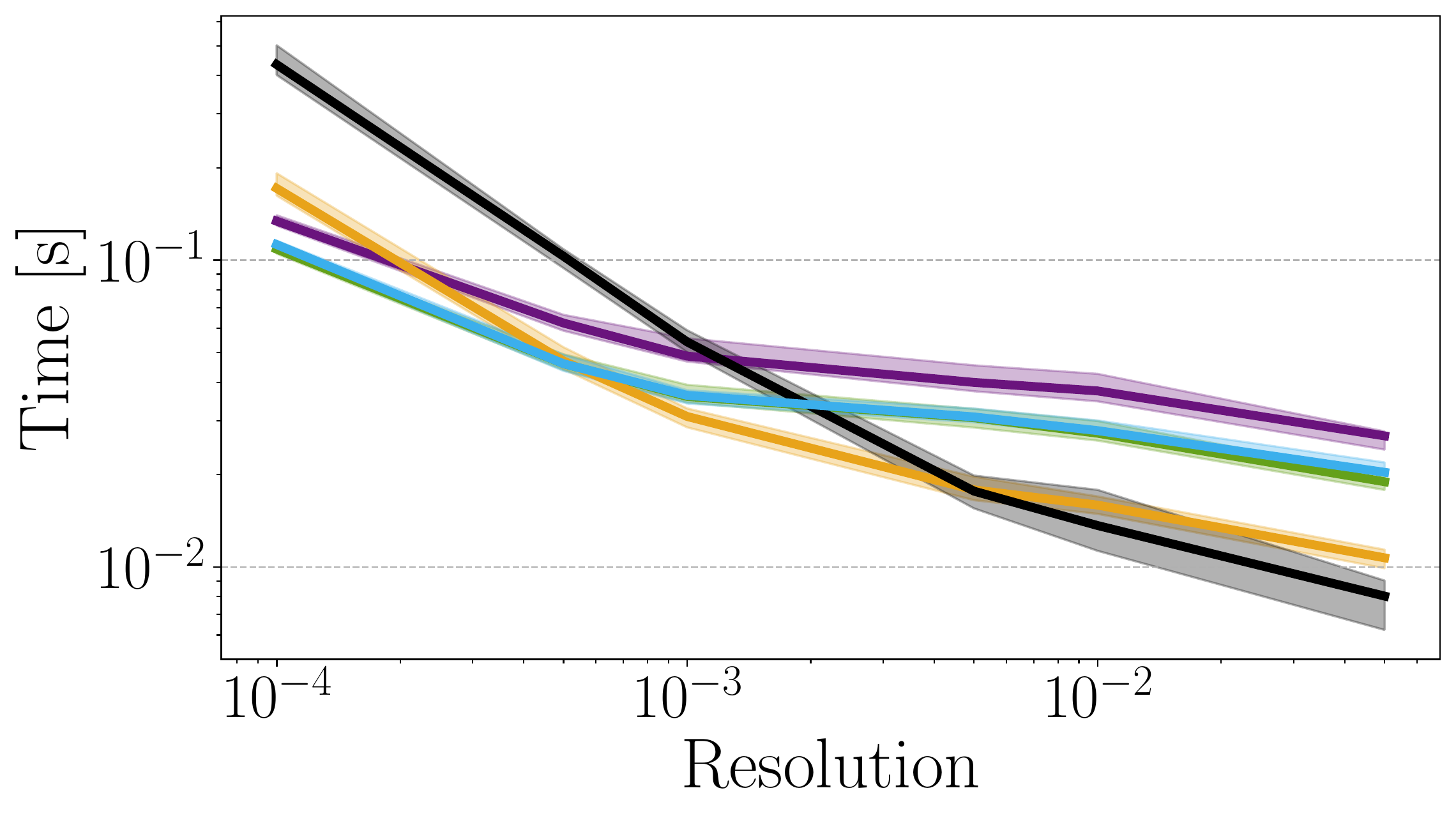}
        \caption{Wall Gap}
    \end{subfigure}%
    \begin{subfigure}[t]{.23\textwidth}
        \centering
        \includegraphics[width=0.97\linewidth]{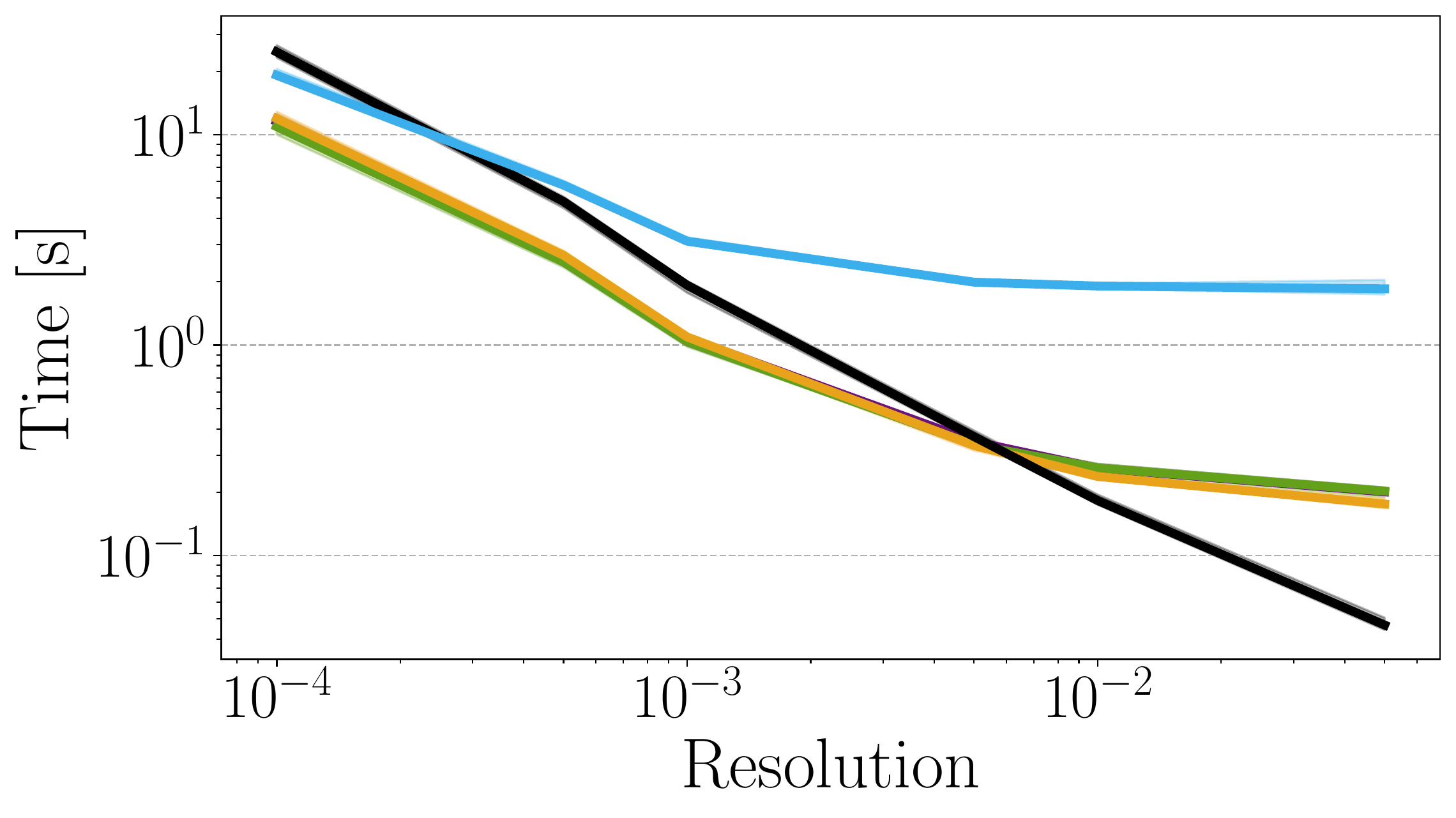}
        \caption{Handover}
    \end{subfigure}
    \centering
    \\[0.5em]
    \begin{subfigure}[b]{1.0\linewidth}%
        \centering
        \begin{tikzpicture}
\begin{axis} [
  width=\textwidth,
  height=0.5\textwidth,
  unbounded coords=jump,
  xtick align=inside,
  ytick align=inside,
  anchor=north,
  hide axis,
  xmajorgrids,
  ymajorgrids,
  major grid style={densely dotted, black!20},
  xmin=0,
  xmax=10,
  ymin=0,
  ymax=10,
  xlabel style={font=\footnotesize},
  xticklabel style={font=\footnotesize},
  ylabel style={font=\footnotesize},
  yticklabel style={font=\footnotesize},
  legend style={anchor=south, legend cell align=left, legend columns=3, at={(axis cs:5, 6)}, font=\small}
]
\addlegendimage{espblack, line width = 1.0pt, mark size=1.0pt, mark=square*}
\addlegendentry{RRT-Connect}
\addlegendimage{esplightblue, line width = 1.0pt, mark size=1.0pt, mark=square*}
\addlegendentry{EIT*}
\addlegendimage{empty legend}
\addlegendentry{}
\addlegendimage{espyellow, line width = 1.0pt, mark size=1.0pt, mark=square*}
\addlegendentry{LazyPRM*}
\addlegendimage{esppurple, line width = 1.0pt, mark size=1.0pt, mark=square*}
\addlegendentry{eo-LazyPRM*}
\addlegendimage{espgreen, line width = 1.0pt, mark size=1.0pt, mark=square*}
\addlegendentry{EIRM*}

\end{axis}
\end{tikzpicture}
    \end{subfigure}
    \vspace{-5mm}
    \caption{
    Median time to find the initial solution for 100 runs at different collision checking resolutions.
    The shaded area shows the 95\% nonparametric confidence bounds.
    }
    \label{fig:coll_res}
\end{figure}

We compare the planning times taken for two scenarios with different collision checking resolutions in \cref{fig:coll_res}.
RRT-Connect scales roughly linear with the collision checking resolution, whereas the search based planners scale better.

Note that in these settings, a collision checking resolution bigger than $10^{-3}$ resulted in an intolerably high chance of obtaining an invalid path.

\section{Discussion \& Limitations}
The planners that reuse collision information over multiple planning queries consistently outperform planners that do not in initial solution time, and are of compareable or better solution quality (\cref{tab:results}).
The difference to planners that do not reuse information is most pronounced when the same problem is solved multiple times, as is the case in a TAMP problem, or a long horizon problem that is solved by decomposing the problem in smaller subproblems (\cref{fig:multiquery_times}).
The approach of decomposing and reusing collision information is more useful if many queries are expected in the same or a similar problem, and the path planning problem is not trivial.

As visible in the collision checking resolution ablation in \cref{ssec:coll_ablation}, the approach presented here shows most improvement in scenarios where the collision checking is relatively time-intensive, as is typically the case in real robotics examples.
While there is an advantage to using the presented approach in simple scenarios as well, it is not as pronounced.

Following the analysis in \cite{Kleinbort2020} the nearest-neighbour computation starts to dominate the planning time once the path is optimized and the number of samples increases.
A possible approach to decrease this part of the planning time as well would be the implementation of caching nearest neighbour computations.

We believe that the presented approach would benefit from a rigorous analysis of when and how to promote valid edges to earlier batches, as already discussed for EIRM* in \cite{hartmann_arxiv22}.
A possible approach to achieve this might be similar to \cite{ichter2018learning}, i.e., online learning the central features of the configuration space, and promoting those to the first batch of samples.

\section{Conclusion}

We presented an approach for algorithms to reuse information in sequential manipulation planning by decomposing collision checking into information that is invariant to changes in the environment, and a part that is not.
This allows efficient reuse of previously computed information over multiple planning queries as they occur in manipulation planning, leading to path planning times that are up to twice as fast than the current state of the art.

While we presented the decomposition-based approach to reuse effort on the example of collision checks, the idea behind it could be applied more generally on motion-validity and state-validity checks.
For example, force or torque constraints could also be decomposed into independent parts, and parts depending on the specific parametrization of a scene. 

In the future, we want to apply this approach to TAMP scenarios, where reuse of the form that we presented arises naturally during planning in one action sequence, where multiple high-level plans are tested, and multiple different grasp positions need to be evaluated.
We also believe that this approach would be suitable in decomposed multi-robot path planning \cite{22-grothe-ICRA}.

In this work, we only focused on the motion planning part, and both the action sequence, and the grasp positions were given.
In future work, the information that is gained during the motion planning queries could be used to e.g. help compute valid grasp positions by biasing towards grasps that were found to be feasible before.

\bibliographystyle{IEEEtran}
\bibliography{IEEEabrv,bib/bib}

\end{document}